\documentclass[10pt,twocolumn,letterpaper]{article}

\usepackage{dicta}
\usepackage{times}
\usepackage{epsfig}
\usepackage{graphicx}
\usepackage{amsmath}
\usepackage{amssymb}

\usepackage{booktabs}

\usepackage{colortbl} 
\usepackage[table]{xcolor}
\usepackage{enumitem}

\usepackage[pagebackref=true,breaklinks=true,letterpaper=true,colorlinks,bookmarks=false]{hyperref}

\dictafinalcopy 

\def\dictaPaperID{15} 

\ifdictafinal\fi
\begin{document}

Copyright 2025 IEEE. Published in the Digital Image Computing: Techniques and Applications, 2025 (DICTA 2025), 3-5 December 2025 in Adelaide, South Australia, Australia. Personal use of this material is permitted. However, permission to reprint/republish this material for advertising or promotional purposes or for creating new collective works for resale or redistribution to servers or lists, or to reuse any copyrighted component of this work in other works, must be obtained from the IEEE. Contact: Manager, Copyrights and Permissions / IEEE Service Center / 445 Hoes Lane / P.O. Box 1331 / Piscataway, NJ 08855-1331, USA. Telephone: + Intl. 908-562-3966.
\title{Exploring Primitive Visual Measurement Understanding and the Role of Output Format in Learning in Vision-Language Models }

\author{
    Ankit Yadav\textsuperscript{\rm 1} \and
    Lingqiao Liu\textsuperscript{\rm 1} \and
    Yuankai Qi\textsuperscript{\rm 2} \\
    \textsuperscript{\rm 1} The University of Adelaide, Adelaide, Australia\\
    \textsuperscript{\rm 2} Macquarie University, Sydney, Australia\\
    ankit.yadav@adelaide.edu.au, lingqiao.liu@adelaide.edu.au, yuankai.qi@mq.edu.au
}

\author{Ankit Yadav\\
The University of Adelaide\\
Adelaide, Australia\\
{\tt\small ankit.yadav@adelaide.edu.au}
\and
Lingqiao Liu\\
The University of Adelaide\\
Adelaide, Australia\\
{\tt\small lingqiao.liu@adelaide.edu.au}
\and
Yuankai Qi\\
Macquarie University\\
Sydney, Australia\\
{\tt\small yuankai.qi@mq.edu.au}
}

\maketitle

\begin{abstract}
   This work investigates the capabilities of current vision-language models (VLMs) in visual understanding and attribute measurement of primitive shapes using a benchmark focused on controlled 2D shape configurations with variations in spatial positioning, occlusion, rotation, size, and shape attributes such as type, quadrant, center-coordinates, rotation, occlusion status, and color as shown in Figure~\ref{fig:sentence_vs_tuple} and supplementary Figures~S3$\sim$S8\footnote{All Figures and Tables with the prefix 'S' refer to the appendix materials. Code: \url{https://github.com/drkkgy/Exploring-Primitive-Visual-Measurement} }. 
We fine-tune state-of-the-art VLMs (2B$\sim$8B parameters) using Low-Rank Adaptation (LoRA) and validate them on multiple out-of-domain (OD) scenarios from our proposed benchmark.
Our findings reveal that coherent sentence-based outputs outperform tuple formats, particularly in OD scenarios with large domain gaps.
Additionally, we demonstrate that scaling numeric tokens during loss computation enhances numerical approximation capabilities, further improving performance on spatial and measurement tasks. 
These results highlight the importance of output format design, loss scaling strategies, and robust generalization techniques in enhancing the training and fine-tuning of VLMs, particularly for tasks requiring precise spatial approximations and strong OD generalization.
\end{abstract}
\section{Introduction}

The advent of Large Language Models (LLMs), such as GPT-2 \cite{radford2019language}, has led to a transformative shift in natural language processing (NLP), with these models demonstrating remarkable performance on complex reasoning tasks and exhibiting strong generalization capabilities. This progress has paved the way for multi-modal LLMs, such as vision-language models (VLMs), which integrate advanced reasoning with vision understanding to address challenges in tasks like Visual Question Answering (VQA).

\begin{figure}[htb]
    \centering
    \includegraphics[width=\linewidth]{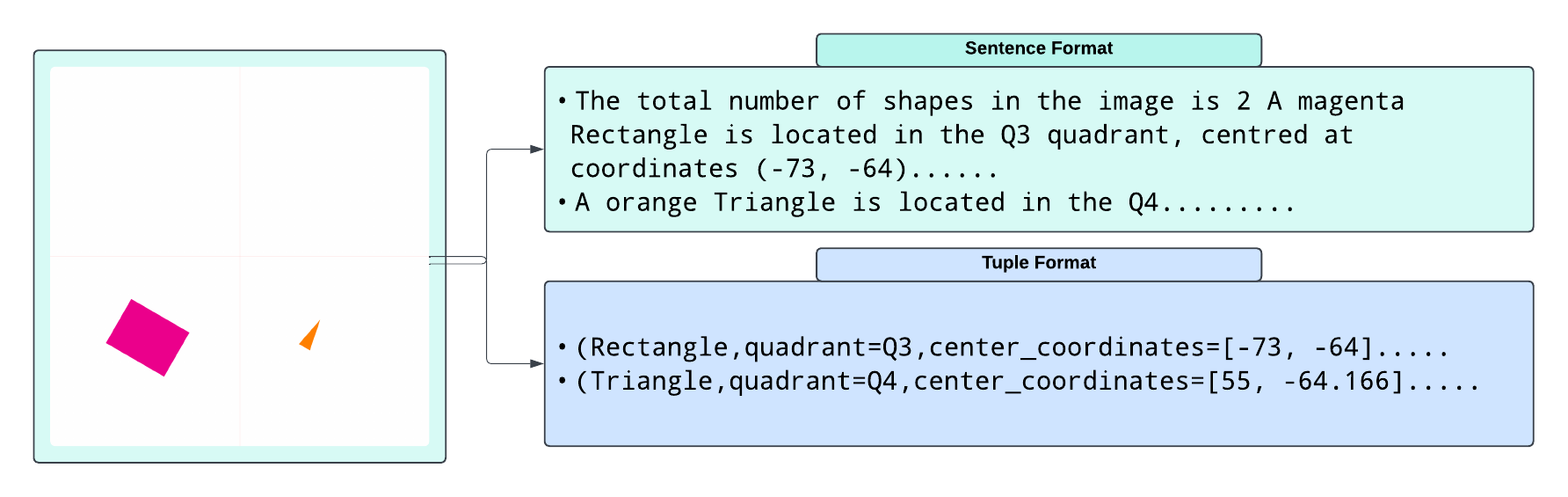}
    \caption{Sentence and Tuple Output Formats used for Fine-tuning.}
    \label{fig:sentence_vs_tuple}
\end{figure}

Significant advancements in this field include models like LLaVA \cite{liu2024visual} and MiniCPM-V \cite{yao2024minicpm}. LLaVA employs a linear projection layer to map vision inputs into the language model's space, while MiniCPM-V uses a re-sampler with attention mechanisms for similar cross-modal projections. These approaches are foundational to several open-source models, including Paligemma \cite{beyer2024paligemma}, Phi-V \cite{abdin2024phi}, and Qwen-VL \cite{bai2023qwen}, which exhibit robust multi-modal capabilities and excel in various VQA tasks \cite{qiao2024prism}.

Despite these successes, an underexplored area is the ability of VLMs to perform precise spatial measurement approximations, such as detecting shapes, determining their attributes, and analyzing their spatial interactions. 
To mitigate this gap, we introduce a synthetic benchmark (Section~\ref{Section-3.2}) to evaluate VLM performance on tasks such as shape counting, shape identification, and center coordinate approximation. 
We also validate these findings on a real-world plant phenotyping dataset \cite{minervini2016finely}, where the goal is to count leaves and predict their center coordinates. We intentionally keep this benchmark simple to first assess the model’s capabilities in fundamental settings, and we plan to extend the benchmark to more complex challenges in future work.

For this study, we fine-tune popular medium and small-sized open-source VLMs on the benchmark dataset and assess their performance in out-of-distribution (OD) scenarios. The analysis evaluates the spatial understanding and measurement capabilities of VLMs and examines whether they can generalize their reasoning abilities to achieve robust spatial comprehension.

\textbf{Our main contributions are summarized as follows:}
\begin{itemize}[itemsep=0.3ex, topsep=0.5ex]
    \item We construct a novel benchmark to evaluate object attribute detection, spatial understanding, and measurement capabilities of VLMs.
    \item We comprehensively evaluate popular open-source VLMs on the benchmark and a real-world dataset.
    \item We investigate into the impact of different output formats on fine-tuning performance.
    \item We introduce loss scaling strategies to improve numerical approximations for VLMs.
    \item We design effective strategies for evaluating compositional outputs of shapes and their attributes.
\end{itemize}

This work discusses the advancements in Vision-Language Models, presents the methodology for creating the benchmark dataset, and details the experiments and results, highlighting key insights into VLM capabilities with potential applications in domains such as autonomous navigation, robotics, and computer vision.

\section{Related Work}
Recent advancements in the reasoning capabilities of Large Language Models (LLMs) have driven the development of Vision-Language Models (VLMs), which combine textual and visual domains to enable advanced visual reasoning. This section reviews key areas of research relevant to VLMs, including Visual Question Answering (VQA), compositional and logical reasoning, and image captioning, while highlighting their limitations and gaps.

\subsection{Visual Question Answering (VQA)}

Visual Question Answering (VQA) is a prominent benchmark for evaluating VLMs. It involves image-question pairs where the question may directly relate to the image (e.g., identifying objects or attributes) or address abstract concepts inferred from the image \cite{sinha2024guiding}. VLMs \cite{yao2024minicpm}\cite{yang2024qwen2} have demonstrated superior zero-shot inference capabilities by leveraging external knowledge beyond the image content to generate coherent and robust responses. However, challenges remain in achieving fine-grained reasoning and spatial granularity, as prior work \cite{shi2024needlargervisionmodels} shows that scaled images often improve performance, making it difficult to determine the optimal scale for detection tasks. This work further investigates VLMs' ability to detect shapes and attributes across OD settings to better understand these challenges. 

\subsection{Compositional and Logical Reasoning in VLMs}

Compositional and logical reasoning are critical for assessing the deeper capabilities of VLMs. Datasets such as GQA (Graph Question Answering) \cite{hudson2019gqa} and CLEVR-X \cite{salewski2020clevr} are specifically designed to test VLMs' ability to perform logical deductions and handle compositional queries.
Another important benchmark, FlowVQA \cite{singh2024flowvqa} evaluates whether VLMs effectively utilize visual inputs in generating responses. Prior studies indicate that VLMs often rely heavily on language priors from the LLM component \cite{kv2020reducing} and sometimes make blind guesses rather than deriving conclusions based on visual data \cite{rahmanzadehgervi2024vision}. 
These findings align closely with our focus on investigating the measurement approximation capabilities of VLMs, particularly in tasks requiring spatial reasoning and attribute detection. They also suggest that VLMs rely less on perceptual clues, which we will examine by testing their detection of shapes and attributes in diverse OD settings.

\subsection{Image Captioning}

Image captioning serves as another important task to evaluate the descriptive abilities of VLMs. 
This involves generating coherent textual descriptions based on visual inputs, a task where recent VLMs with strong pre-trained LLM backbones have significantly improved performance. 
Early models like Flamingo \cite{alayrac2022flamingo} used vision and cross-modal adapters, while recent advancements like MiniCPM-V \cite{yao2024minicpm} incorporate adaptive visual encoding and robust cross-modal resamplers. 
These models leverage state-of-the-art LLMs such as LLama3 \cite{dubey2024llama} and Qwen2 \cite{yang2024qwen2} to enhance caption generation quality and scene understanding. 
However, despite these advancements, the ability of VLMs to understand complex scenes with intricate object interactions and to perform fine-grained spatial measurements is an area that remains underexplored, as captions typically focus on high-level descriptions rather than detailed spatial reasoning.
\section{Methodology}

Vision-Language Models (VLMs) have shown impressive performance in object detection and attribute description. However, existing benchmarks focus mainly on coarse detection and high-level interactions, leaving fine-grained measurement tasks underexplored.

This work evaluates VLMs' ability to detect shape attributes and approximate spatial measurements, specifically estimating center coordinates and rotational angles of geometric shapes. To support this, we introduce a benchmark with multiple out-of-domain (OD) test sets designed to assess spatial reasoning and attribute detection.

In the following section, we present the benchmark dataset, evaluation metrics, and fine-tuning strategies, along with a novel metric, Structured Attribute Matching Accuracy (SAMA), to provide deeper insights into VLM performance across diverse OD settings.

\subsection{Pre-trained Model Selection }
\label{Section-3.1}

Our study focus on the following popular small and medium open-source Vision Language Models, which are widely adopted by researchers and small businesses thanks to its affordable cost: 


\noindent\textbf{Evaluated Models:}
\begin{itemize}[noitemsep,leftmargin=1.5em]
  \item MiniCPM-V-2.6(8B)~\cite{yao2024minicpm}
  \item MiniCPM-V-2.5(8B)~\cite{yao2024minicpm}
  \item Qwen2-VL-2B-Instruct(2B)~\cite{wang2024qwen2}
  \item Qwen2-VL-7B-Instruct(7B)~\cite{wang2024qwen2}
  \item Phi-3.5-Vision-Instruct(4.5B)~\cite{abdin2024phi}
  \item Paligemma-3B-PT-224(3B)~\cite{beyer2024paligemma}
\end{itemize}


The medium models like Qwen2-VL-7B and MiniCPM-V models have 7B and 8B parameters while the smaller VLMs are in the range of 2B to 4.5B parameters. We consider models of different sizes to study the impact of model size on measurement capabilities and focus on open-source models due to their accessibility, transparency, and reproducibility.  
For diversity, we consider both the MiniCPM-V versions, the MiniCPM-V-2.5 with the LLama3-8B and MiniCPM-V-2.6 with the Qwen-2 as the LLM backbone.

\subsection{Proposed Benchmark}
\label{Section-3.2}

\subsubsection*{Benchmark Configuration}

This benchmark dataset evaluates Vision-Language Models (VLMs) on object attribute detection, spatial understanding, and measurement tasks using simple 2D shapes. The dataset assesses VLM performance under varying OD conditions of shape composition, occlusion, color variation, and rotation. 

The dataset comprises:
\begin{itemize}
    \item \textbf{Training Set}: 20,000 samples with diverse shape configurations and attributes.
    \item \textbf{Evaluation Set}: 1,000   samples, sharing the training set’s configuration (detailed below) but unique examples ensured using MD5 hashing\cite{rivest1992rfc1321}.
    \item \textbf{Test Set}: equally divided into five configurations of 200 samples each, as shown in Table~\ref{tab:od-scenarios}.
\end{itemize}

\begin{table}[t]
  \centering
  \small
  \resizebox{\columnwidth}{!}{%
    \begin{tabular}{@{}lccccc@{}}
      \toprule
      \textbf{OD Test Set}    & \textbf{Shapes} & \textbf{Occlusion} & \textbf{New Rotation} & \textbf{Size} & \textbf{Spatial Layout} \\
      \midrule
      OD Composition       & 5–6             & 5–6 overlaps       & 45° \& 72°           & ×1           & novel                   \\
      OD Spatial Awareness & 5–6             & as training        & as training          & ×1           & novel                   \\
      OD Occlusion         & as training     & 4–5 overlaps       & as training          & ×1           & as training             \\
      OD Rotation          & as training     & as training        & 45° \& 72°           & ×1           & as training             \\
      OD Size              & as training     & as training        & as training          & ×2           & as training             \\
      \bottomrule
    \end{tabular}
  }
  \caption{Out-of-Domain Test Sets}
  \label{tab:od-scenarios}
\end{table}

\paragraph{Training Set Configuration:}

The training set features 2$\sim$4 shapes from $\{$Circle, Rectangle, Ellipse, Triangle, Square$\}$ per image, with six possible colors $\{$Orange, Red, Blue, Green, Yellow, Magenta$\}$.
Rotations of 0$^\circ$, 15$^\circ$, and 30$^\circ$ are applied randomly while ensuring rotational uniqueness (e.g., 15$^\circ$ is not treated as 105$^\circ$ or other symmetric equivalent shapes). Up to three shapes are allowed to overlap per image. MD5 hashing \cite{rivest1992rfc1321} ensures all configurations are unique, eliminating duplicates. Further information on the dataset generation process can be found in the supplementary Section~A2.1.

\paragraph{Output Formats:}

We fine-tune the models on the training set of the benchmark and evaluate their performance using two distinct output formats
(Figure~\ref{fig:sentence_vs_tuple}).

\textbf{Sentence Format:} Shapes and their attributes are organized into structured, coherent natural language sentences, aligning with the pretraining objectives and stylistic preferences of large language models (LLMs).

\textbf{Tuple Format:} Shapes and their attributes are represented in a structured tuple format, similar to JSON outputs commonly used for structured data in deep learning workflows.

\paragraph{Applied Evaluation: Plant Phenotyping Dataset}

To validate our findings in real-world scenarios, we employ the Plant Phenotyping dataset as described in \cite{minervini2016finely}. This dataset consists of images of plants captured at various developmental stages, annotated with both leaf counts and corresponding bounding boxes. Using these annotations, we compute the centroids of the bounding boxes to approximate the center points of the leaves (Supplementary Figure~S17). 

The models are fine-tuned to predict both the number of leaves and their approximate center coordinates. Additionally, to thoroughly evaluate the models, we construct a series of out-of-domain (OD) test sets by combining different subgroups of the dataset. These configurations range from easy to challenging, enabling a comprehensive assessment of the models’ generalization capabilities. Further details are provided in the supplementary section and Figure~S16.

\subsection{Fine-Tuning Setup}

For fine-tuning, we employ Low-Rank Adaptation (LoRA) across the attention layers of both the language model (LLM) and vision processing modules within each model architecture, facilitating adaptation to the challenging task. Ablation studies, as detailed in Supplementary Table~S1, further validate this approach. Additionally, LoRA is applied to the attention components within the resampler or bottleneck layers. Below, we outline the specific layers fine-tuned for each model.

\textbf{MiniCPM-V-2.6:} We fine-tune the self-attention layers across both the language and vision components. Specifically, we adapt the key, query, value, and output-projection layers in the self-attention modules. Additionally, we apply LoRA to the key-value layers within the re-sampler modules, ensuring that the cross-modal attention mechanisms are optimized during fine-tuning.

\textbf{MiniCPM-V-2.5:} In this model, LoRA is applied to the self-attention layers within the LLM, targeting the key, query, value, and output-projection layers of the attention components. Similar to MiniCPM-V-2.6, the key-value layers within resamplers are fine-tuned. Furthermore, in the vision processing module (VPM), we fine-tune all encoder attention layers, focusing on the key, query, value, and output components in the self-attention modules.

\textbf{Phi-3.5-V:} We finetune attention layers across both the language and vision encoders. LoRA is applied to the attention layers in the vision embedding module including key, query, value, and output projection. For each layer in the LLM, we adapt the unified query-key-value and output in the self-attention components and the up-projection and down-projection in the MLP blocks.

\textbf{Paligemma-3b:} In Paligemma, fine-tuning is focused on both the vision tower and language model components. Within the vision tower, LoRA is applied to the self-attention layers, including key, query, value, and output projection, across all encoder layers. In the LLM layers, LoRA is applied to the key, query, value, and output-projection layers of the attention component. Additionally, LoRA is applied to the multi-layer perceptron (MLP) components in the encoder, specifically the first and second fully connected layers, as well as the gating, up, and down projection layers.

\textbf{Qwen2-VL:} For Qwen2-VL, fine-tuning is conducted across both the visual and language modules. In the visual component, LoRA is applied to the unified query-key-value and output projections. For the language model, we adapt each self-attention layer’s key, query, value, and output projections and the up and down projections in the MLP components across all layers.


\paragraph{LoRA Configuration}

For our benchmark dataset, all models are configured with a rank of 64, an alpha value of 64, and a dropout rate of 0.05. For the applied experiments with the Plant Phenotyping dataset, the MiniCPM-V models are configured with a rank of 8, an alpha value of 16, and a dropout rate of 0.15. The reduced rank for this dataset is motivated by its smaller size, aiming to prevent overfitting and optimize performance.

\subsection{Structured Attribute Matching Accuracy (SAMA) Based on the Jonker-Volgenant Algorithm}
\label{Section-3.4}

\begin{figure}[htbp]
    \centering
    \includegraphics[width=1\linewidth]{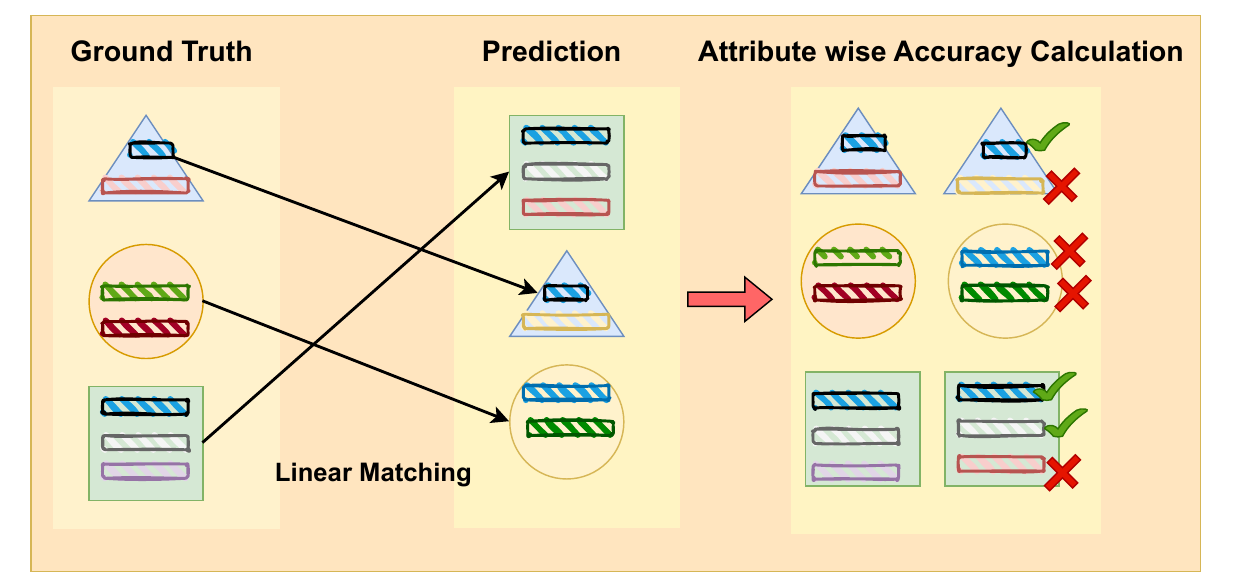}
    \caption{Structured Attribute Matching Accuracy Calculation for a single data point.}
    \label{fig:custom_accuracy_explanation}
\end{figure}

We propose a \textbf{Custom Accuracy} metric leveraging the Jonker-Volgenant (JV) algorithm as described in \cite{crouse2016implementing} to address attribute matching challenges inherent in our multi-attribute compositional image prediction task (detailed below). We assume model outputs are properly structured to enable the JV algorithm’s edit-distance cost function and regex-based attribute extraction.
In our experiment, we use the VLMs to detect different shapes and their attributes in the image like color, occlusions etc. 
Each prediction may contain multiple shapes and corresponding attributes presented in arbitrary order relative to the ground truth. This discrepancy introduces a Linear Assignment Problem (LAP) where predicted attributes and shapes require optimal matching with ground truth data. 
Inspired by \cite{carion2020end},  we employ the JV algorithm, utilizing edit distance algorithm \cite{hyyro2001explaining} as the cost function to align predicted shapes and attributes with those in the ground truth. We chose the edit distance-based cost function in the JV algorithm as it is sensitive to slight changes hence it enhances the robustness of our Custom Accuracy metric.
Once matched, each attribute (shape, color, occlusion status, etc.) is extracted via regex\footnote{Regex is a powerful tool for pattern matching and string manipulation in programming. For Python, see \url{https://docs.python.org/3/library/re.html}.}, allowing for a detailed evaluation of correctly predicted attributes within each pair (Figure~\ref{fig:custom_accuracy_explanation}). 
We calculate the accuracy for each attribute pair within a prediction, then compute the average accuracy across all predictions, which we term \textbf{Structured Attribute Matching Accuracy (SAMA)}. 
For continuous attributes, such as center coordinates and rotation angle, we apply the same matching process, calculating the Root Mean Squared Error (RMSE) for each matched pair and similarly taking the average. 

\subsection{Precision and Recall-Based Evaluation for Attribute Detection}
\label{Section-3.5}

To evaluate the precision and recall of attribute detection, we measure the frequencies of attributes being correctly identified. Specifically, we compare the ground truth (GT) and predicted (PT) values for each attribute by analyzing the frequency distribution of each class for the given attribute.

For instance, consider the attribute \emph{shape type}, which consists of six classes\texttt{[Circle, Rectangle, Ellipse, Triangle, Square, NA]}. It is important to note that the \texttt{NA} class is not included in the evaluation calculations and serves only as a placeholder to ensure that the GT and PT vectors are of the same size. Given a ground truth set $GT = [\text{circle}, \text{circle}, \text{triangle}]$ and a predicted set $PT = [\text{square}, \text{triangle}, \text{circle}]$, the corresponding frequency vectors are
\begin{equation}
\begin{aligned}
  \mathrm{GT} &= [2,0,0,1,0,0],\;\;
  \mathrm{PT} = [1,0,0,1,1,0],\\[-0.4ex]
  \mathrm{correct} &= \min(\mathrm{GT},\mathrm{PT})
                   = [1,0,0,1,0,0],\\[-0.4ex]
  &\begin{aligned}
    \text{total\_correct} &= \sum \mathrm{correct} = 2,\\
    \text{true\_total}    &= \sum \mathrm{GT}      = 3 \\
    \text{pred\_total}    &= \sum \mathrm{PT}      = 3 
  \end{aligned}
  \quad
\end{aligned}
\label{eq:all-counts}
\end{equation}
\begin{equation}
\begin{aligned}
  \mathrm{Precision} &= 
    \dfrac{\text{total\_correct}}{\text{pred\_total}}
      \quad(\text{pred\_total}>0),\\[-0.4ex]
    &\quad 0 \quad(\text{otherwise});\\[1ex]
  \mathrm{Recall}    &= 
    \dfrac{\text{total\_correct}}{\text{true\_total}}
      \quad(\text{true\_total}>0),\\[-0.4ex]
    &\quad 0 \quad(\text{otherwise}).
\end{aligned}
\label{eq:pr}
\end{equation}

\subsection{Loss Scaling}

MiniCPM-V2.5 and MiniCPM-V2.6 adopt distinct tokenization strategies for numeric values. MiniCPM-V2.5 uses predefined tokens for values between 1 and 1000, while MiniCPM-V2.6 limits this range to 1 to 10. Numeric values outside these ranges are mapped to a single predefined token in both models. Other models, such as Qwen2-VL, employ similar tokenization strategies.

As part of our methodology, scaling is applied exclusively to the loss values associated with these predefined numeric tokens as depicted in  Figure~S18 to investigate their impact on the numeric approximation performance of VLMs and to evaluate the effectiveness of token scaling in improving overall performance.


\begin{table}[t]
    \centering
    \resizebox{\columnwidth}{!}{%
    \begin{tabular}{lcccccc}
        \toprule
        \textbf{Model} & \textbf{OD Comp.} & \textbf{OD Occl.} & \textbf{OD Rot.} & \textbf{OD Spatial} & \textbf{Test Set} & \textbf{OD Size} \\
        \midrule
        \multicolumn{7}{c}{\textit{Tuple-Version Results}} \\
        \midrule
        MiniCPM-V2.6 & 0.504 & 0.654 & 0.649 & 0.517 & 0.654 & \textbf{0.605} \\
        MiniCPM-2.5 & 0.395 & 0.574 & 0.532 & 0.431 & 0.573 & 0.512 \\
        Qwen-VL 7B & 0.110 & 0.065 & 0.062 & 0.107 & 0.063 & 0.059 \\
        Qwen-VL 2B & 0.048 & 0.013 & 0.007 & 0.045 & 0.010 & 0.008 \\
        Phi-V & 0.027 & 0.013 & 0.012 & 0.025 & 0.012 & 0.012 \\
        Paligemma3B & 0.102 & 0.054 & 0.056 & 0.087 & 0.048 & 0.061 \\
        \midrule
        \multicolumn{7}{c}{\textit{Sentence-version Results}} \\
        \midrule
        MiniCPM-V2.6  & \textbf{0.547} & \textbf{0.699} & \textbf{0.688} & \textbf{0.541} & \textbf{0.688} & 0.600 \\
        MiniCPM-2.5 & 0.498 & 0.666 & 0.643 & 0.493 & 0.659 & 0.588 \\
        Qwen-VL 7B & 0.132 & 0.080 & 0.079 & 0.122 & 0.081 & 0.077 \\
        Qwen-VL 2B & 0.054 & 0.004 & 0.002 & 0.052 & 0.002 & 0.003 \\
        Phi-V & 0.024 & 0.013 & 0.013 & 0.024 & 0.024 & 0.014 \\
        Paligemma3B & 0.089 & 0.041 & 0.050 & 0.076 & 0.042 & 0.071 \\
        \bottomrule
        \end{tabular}%
        }
    \caption{Accuracy (SAMA) scores for various Vision-Language Models across different out-of-domain test sets. Higher values indicate better performance.}
    \label{tab:SAMA Accuracy}
\end{table}

\begin{table*}[h]
    \centering
    \footnotesize
    
        \begin{tabular}{l*{6}{cc}}
            \toprule
            & \multicolumn{2}{c}{\textbf{OD Comp.}} & \multicolumn{2}{c}{\textbf{OD Occl.}} & \multicolumn{2}{c}{\textbf{OD Rot.}} & \multicolumn{2}{c}{\textbf{OD Spatial}} & \multicolumn{2}{c}{\textbf{Test Set}} & \multicolumn{2}{c}{\textbf{OD Size}} \\
            \cmidrule(lr){2-3} \cmidrule(lr){4-5} \cmidrule(lr){6-7} \cmidrule(lr){8-9} \cmidrule(lr){10-11} \cmidrule(lr){12-13}
            \textbf{Model} & C↓ & R↓ & C↓ & R↓ & C↓ & R↓ & C↓ & R↓ & C↓ & R↓ & C↓ & R↓ \\
            \midrule
            \multicolumn{13}{c}{\textit{Tuple-Version Results}} \\
            \midrule
            MiniCPM-V2.5 & 64.850 & 36.485 & 40.866 & 7.375 & 45.783 & 25.023 & 64.277 & 17.740 & 40 & 8.585 & 104.246 & 7.675 \\
            MiniCPM-V2.6 & 51.642 & 34.813 & 32.274 & 5.575 & 33.712 & 36.095 & 55.373 & 8.425 & 32.628 & 6.200 & 98.098 & 6.500 \\
            Qwen-VL 7B & 66.396 & 36.435 & 61.218 & 8.980 & 64.068 & 41.315 & 70.419 & 8.375 & 66.850 & 8.463 & 148.344 & 8.870 \\
            Qwen-VL 2B & 81.604 & 33.745 & 83.205 & 11.225 & 81.392 & 42.688 & 83.454 & 9.135 & 89.572 & 10.521 & 191.350 & 11.705 \\
            Ph3i-V & 47.529 & 26.014 & 41.614 & 4.755 & 48.920 & 29.775 & 50.531 & 6.125 & 41.636 & 4.807 & 151.941 & 5.080 \\
            Paligemma3b & 75.717 & 22.685 & 90.702 & 7.407 & 94.313 & 24.903 & 86.827 & 6.036 & 89.511 & 7.175 & 197.858 & 10.070 \\
            \midrule
            \multicolumn{13}{c}{\textit{Sentence-version Results}} \\
            \midrule
            MiniCPM-V2.5 [SF] & 36.778 & 16.652 & 28.415 & 5.450 & 26.025 & 20.835 & 38.216 & 6.450 & 22.547 & 5.410 & 84.847 & 6.775 \\
            MiniCPM-V2.6 [SF] & \textbf{34.049} & 16.125 & \textbf{24.240} & 4.250 & \textbf{23.293} & 22.063 & \textbf{34.134} & 4.600 & \textbf{19.772} & 4.570 & \textbf{80.660} & 6.050 \\
            Qwen-VL 7B [SF] & 50.033 & 15.365 & 55.246 & 6.250 & 48.277 & 21.605 & 44.948 & 3.483 & 51.185 & 6.295 & 150.295 & 6.295 \\
            Ph3i-V [SF] & 34.961 & 11.675 & 37.862 & \textbf{4.025} & 32.481 & 18.385 & 36.460 & \textbf{2.885} & 27.963 & \textbf{2.910} & 168.317 & \textbf{4.145} \\
            Paligemma3b [SF] & 65.888 & \textbf{9.431} & 76.489 & 5.777 & 74.011 & \textbf{18.063} & 73.370 & 4.001 & 72.345 & 4.436 & 163.727 & 4.982 \\
            Qwen-VL 2B [SF] & 90.314 & 35.219 & 95.293 & 10.650 & 90.501 & 44.849 & 92.709 & 9.540 & 97.390 & 7.455 & 196.669 & 11.150 \\
            \bottomrule
        \end{tabular}
    
    \caption{Performance comparison of Vision-Language Models on OD test datasets, showing Center (C) and Rotation (R) RMSE. ↓ indicates lower values are better.}
    \label{tab:center_coord_rotation_rmse}
\end{table*}

\begin{table}[h]
   \centering
   \resizebox{\columnwidth}{!}{
   \begin{tabular}{l|c|c|c|c}
       \toprule
       & \multicolumn{1}{c|}{\textbf{Test Set}} & \multicolumn{1}{c|}{\textbf{Train Set}} & \multicolumn{1}{c|}{\textbf{OD Test Set}} & \multicolumn{1}{c}{\textbf{Strong OD Test Set}} \\
       \textbf{Model} & Count↓ & Count↓ & Count↓ & Count↓ \\
       \midrule
       Plant\_sentence & \textbf{1.19} & \textbf{1.78} & \textbf{1.59} & \textbf{3.01} \\
       Plant\_Tuple & 1.31 & 1.99 & 1.65 & 6.28 \\
       \bottomrule
   \end{tabular}
   }
   \caption{Performance comparison between sentence and tuple formats on the Plant Phenotyping dataset. The gap widens in favour of sentence format as out-of-distribution difficulty increases. ↓: lower is better. Count: Count RMSE.}
   \label{tab:model_performance_plant_dataset}
\end{table}

\begin{table}[h]
   \centering
   \resizebox{\columnwidth}{!}{
   \begin{tabular}{lcccccc}
       \toprule
       \textbf{Model} & \textbf{OD Comp.} & \textbf{OD Occl.} & \textbf{OD Rot.} & \textbf{OD Spatial} & \textbf{Test Set} & \textbf{OD Size} \\
       \midrule
       \multicolumn{7}{c}{\textit{Tuple-Version Results}} \\
       \midrule
       MiniCPM-V2.6 & 51.60 & 32.30 & 33.70 & 55.40 & 32.60 & 98.10 \\
       \midrule
       \multicolumn{7}{c}{\textit{Sentence-version Results}} \\
       \midrule
       MiniCPM-V2.6 [SF] & 34.00 & 24.20 & 23.30 & 34.10 & 19.80 & 80.66 \\
       MiniCPM-V2.6 SW 3.5 & 36.53 & 22.52 & 22.02 & 36.02 & \textbf{18.21} & \textbf{79.38} \\
       MiniCPM-V2.6 SW 2.0 & \textbf{31.81} & \textbf{21.53} & 22.94 & 32.44 & 18.45 & 82.39 \\
       MiniCPM-V2.6 SW 2.5 & 34.66 & 21.67 & 22.46 & 32.77 & 18.36 & 82.37 \\
       MiniCPM-V2.6 SW 1.5 & 35.20 & 22.45 & \textbf{21.85} & \textbf{31.56} & 18.89 & 80.77 \\
       \bottomrule
   \end{tabular}
   }
   \caption{Center Coordinate RMSE comparison across different model variants on OD test datasets. SW indicates sentences weighted with different scale factors. ↓ indicates lower values are better.}
   \label{tab:accuracy_scores for scaled loss}
\end{table}

\begin{table*}[t]
    \centering
    \tiny
    
        \begin{tabular}{l*{6}{ccc}}
            \toprule
            & \multicolumn{3}{c}{\textbf{OD Comp.}} & \multicolumn{3}{c}{\textbf{OD Occl.}} & \multicolumn{3}{c}{\textbf{OD Rot.}} & \multicolumn{3}{c}{\textbf{OD Spatial}} & \multicolumn{3}{c}{\textbf{Test Set}} & \multicolumn{3}{c}{\textbf{OD Size}} \\
            \cmidrule(lr){2-4} \cmidrule(lr){5-7} \cmidrule(lr){8-10} \cmidrule(lr){11-13} \cmidrule(lr){14-16} \cmidrule(lr){17-19}
            \textbf{Model} & P↑ & R↑ & \cellcolor{gray!20}F1↑ & P↑ & R↑ & \cellcolor{gray!20}F1↑ & P↑ & R↑ & \cellcolor{gray!20}F1↑ & P↑ & R↑ & \cellcolor{gray!20}F1↑ & P↑ & R↑ & \cellcolor{gray!20}F1↑ & P↑ & R↑ & \cellcolor{gray!20}F1↑ \\
            \midrule
            MiniCPM-V2.5(8B)& 0.803 & 0.646 & \cellcolor{gray!20}0.714 & 0.781 & 0.775 & \cellcolor{gray!20}0.777 & 0.770 & 0.767 & \cellcolor{gray!20}0.766 & 0.799 & \textbf{0.653} & \cellcolor{gray!20}0.717 & 0.852 & 0.843 & \cellcolor{gray!20}0.846 & 0.693 & 0.767 & \cellcolor{gray!20}0.723 \\
            MiniCPM-V2.5(8B) [SF] & \textbf{0.815} & 0.645 & \cellcolor{gray!20}\textbf{0.718} & \textbf{0.798} & \textbf{0.787} & \cellcolor{gray!20}\textbf{0.791} & \textbf{0.793} & \textbf{0.777} & \cellcolor{gray!20}\textbf{0.782} & \textbf{0.816} & 0.651 & \cellcolor{gray!20}\textbf{0.722} & \textbf{0.868} & \textbf{0.857} & \cellcolor{gray!20}\textbf{0.861} & \textbf{0.731} & \textbf{0.783} & \cellcolor{gray!20}\textbf{0.752} \\
            \midrule
            MiniCPM-V2.6(8B) & \textbf{0.816} & \textbf{0.686} & \cellcolor{gray!20}\textbf{0.742} & 0.808 & 0.803 & \cellcolor{gray!20}0.804 & \textbf{0.817} & 0.805 & \cellcolor{gray!20}0.809 & 0.815 & \textbf{0.688} & \cellcolor{gray!20}\textbf{0.744} & 0.884 & 0.875 & \cellcolor{gray!20}0.878 & \textbf{0.769} & 0.785 & \cellcolor{gray!20}\textbf{0.775} \\
            MiniCPM-V2.6(8B) [SF] & 0.814 & 0.672 & \cellcolor{gray!20}0.733 & \textbf{0.816} & \textbf{0.813} & \cellcolor{gray!20}\textbf{0.814} & 0.815 & \textbf{0.807} & \cellcolor{gray!20}\textbf{0.810} & \textbf{0.819} & 0.674 & \cellcolor{gray!20}0.738 & \textbf{0.884} & \textbf{0.878} & \cellcolor{gray!20}\textbf{0.880} & 0.727 & \textbf{0.794} & \cellcolor{gray!20}0.755 \\
            \midrule
            Qwen-VL 7B & 0.715 & 0.567 & \cellcolor{gray!20}0.611 & 0.626 & \textbf{0.504} & \cellcolor{gray!20}\textbf{0.541} & 0.629 & \textbf{0.513} & \cellcolor{gray!20}\textbf{0.549} & 0.714 & 0.562 & \cellcolor{gray!20}0.606 & 0.629 & \textbf{0.513} & \cellcolor{gray!20}\textbf{0.549} & 0.609 & \textbf{0.494} & \cellcolor{gray!20}\textbf{0.529} \\
            Qwen-VL 7B [SF] & \textbf{0.751} & \textbf{0.577} & \cellcolor{gray!20}\textbf{0.637} & \textbf{0.642} & 0.491 & \cellcolor{gray!20}0.540 & \textbf{0.632} & 0.483 & \cellcolor{gray!20}0.532 & \textbf{0.760} & \textbf{0.564} & \cellcolor{gray!20}\textbf{0.630} & \textbf{0.651} & 0.497 & \cellcolor{gray!20}0.547 & \textbf{0.625} & 0.466 & \cellcolor{gray!20}0.516 \\
            \midrule
            Qwen-VL 2B & \textbf{0.669} & \textbf{0.512} & \cellcolor{gray!20}\textbf{0.553} & \textbf{0.578} & \textbf{0.471} & \cellcolor{gray!20}\textbf{0.502} & \textbf{0.582} & \textbf{0.474} & \cellcolor{gray!20}\textbf{0.506} & \textbf{0.660} & \textbf{0.503} & \cellcolor{gray!20}\textbf{0.543} & \textbf{0.600} & \textbf{0.491} & \cellcolor{gray!20}\textbf{0.523} & \textbf{0.578} & \textbf{0.461} & \cellcolor{gray!20}\textbf{0.493} \\
            Qwen-VL 2B [SF] & 0.654 & 0.484 & \cellcolor{gray!20}0.540 & 0.576 & 0.452 & \cellcolor{gray!20}0.491 & 0.561 & 0.455 & \cellcolor{gray!20}0.490 & 0.648 & 0.476 & \cellcolor{gray!20}0.534 & 0.589 & 0.466 & \cellcolor{gray!20}0.505 & 0.546 & 0.430 & \cellcolor{gray!20}0.467 \\
            \midrule
            Phi3-V(4.2B) & 0.447 & \textbf{0.784} & \cellcolor{gray!20}0.489 & 0.448 & 0.824 & \cellcolor{gray!20}0.481 & 0.449 & \textbf{0.827} & \cellcolor{gray!20}0.485 & 0.440 & \textbf{0.796} & \cellcolor{gray!20}0.482 & 0.474 & 0.893 & \cellcolor{gray!20}0.510 & 0.409 & \textbf{0.800} & \cellcolor{gray!20}0.439 \\
            Phi3-V(4.2B) [SF] & \textbf{0.497} & 0.771 & \cellcolor{gray!20}\textbf{0.509} & \textbf{0.480} & \textbf{0.828} & \cellcolor{gray!20}\textbf{0.501} & \textbf{0.491} & 0.826 & \cellcolor{gray!20}\textbf{0.511} & \textbf{0.486} & 0.775 & \cellcolor{gray!20}\textbf{0.491} & \textbf{0.519} & \textbf{0.894} & \cellcolor{gray!20}\textbf{0.540} & \textbf{0.460} & 0.799 & \cellcolor{gray!20}\textbf{0.466} \\
            \midrule
            Paligemma3b(2.92B) & 0.626 & \textbf{0.637} & \cellcolor{gray!20}\textbf{0.618} & \textbf{0.631} & \textbf{0.628} & \cellcolor{gray!20}\textbf{0.579} & \textbf{0.646} & \textbf{0.643} & \cellcolor{gray!20}\textbf{0.585} & 0.635 & \textbf{0.637} & \cellcolor{gray!20}\textbf{0.623} & \textbf{0.647} & \textbf{0.644} & \cellcolor{gray!20}\textbf{0.591} & \textbf{0.600} & \textbf{0.626} & \cellcolor{gray!20}\textbf{0.554} \\
            Paligemma3b(2.92B) [SF] & \textbf{0.662} & 0.614 & \cellcolor{gray!20}0.602 & 0.574 & 0.578 & \cellcolor{gray!20}0.512 & 0.578 & 0.583 & \cellcolor{gray!20}0.514 & \textbf{0.671} & 0.608 & \cellcolor{gray!20}0.601 & 0.580 & 0.581 & \cellcolor{gray!20}0.515 & 0.574 & 0.606 & \cellcolor{gray!20}0.517 \\
            \bottomrule
        \end{tabular}

    \caption{Performance comparison of Vision-Language Models on OD test datasets, showing Precision (P), Recall (R), and F1-score. $\uparrow$ indicates higher values are better.  'SF' denotes the sentence format used}
    \label{tab:precission-recall-f1-score}
\end{table*}

\section{Experiments and Results}
\label{exp:experiment_and_result}
\subsection{Implementation Details and Metrics}

Each experiment is fine-tuned on the benchmark training dataset, evaluated on the evaluation/test dataset, and tested on five out-of-domain (OD) test sets.
Experiments annotated with the tag \texttt{`SF`} are fine-tuned using the Sentence output format, while those without the tag are fine-tuned using the Tuple output format, unless otherwise specified. 

Each fine-tuning experiment employs the following hyperparameters: a learning rate of $1 \times 10^{-6}$ with a cosine scheduler and a warmup ratio of 0.01. Fine-tuning is performed for 10,000 steps using the AdamW optimizer, with $\beta_2$ set to 0.95 and weight decay set to 0.1. For the plant phenotyping dataset \cite{minervini2016finely} used in the applied evaluation experiment, fine-tuning is conducted for only 4,000 steps due to the limited dataset size. 
Experiments are conducted on MiniCPM-V, Phi-3-V, and Qwen-2-VL models (2B and 7B versions) using 2× Nvidia A100 GPUs, and Paligemma models on 2× Nvidia RTX 4060 Ti GPUs due to limited resource availability. Each GPU operates with a batch size of 1, with gradient accumulation set to 8, yielding an effective batch size of 16. We use \texttt{bf16} precision for fine-tuning. These fine-tuning settings are borrowed from \cite{yao2024minicpm}.

\noindent\textbf{Note:} To address the limited size of the Plant Phenotyping dataset and achieve faster convergence, we adopt a consistent prompt template: \texttt{"For the given plant image predict the number of leaves and the center of the leaves"}.

We evaluate the quality of the predicted outputs using SAMA (Section~\ref{Section-3.4}) to measure prediction accuracy, alongside Precision, Recall, and F1 metrics, as outlined in Section~\ref{Section-3.5}. For center coordinate and rotation angle predictions, Root Mean Squared Error (RMSE) is calculated after performing linear matching. On the Plant dataset, RMSE is used to evaluate both leaf count and leaf center coordinates. For linear matching in this dataset, RMSE serves as the cost function instead of edit distance, with linear matching applied prior to calculating RMSE for the center coordinates. We only fine-tune MiniCPM-V2.6 for this dataset.

\subsection{Quantitative Results}

This section presents the quantitative findings, as shown in Tables~\ref{tab:SAMA Accuracy} to~\ref{tab:precission-recall-f1-score}, analyzing the performance of models fine-tuned using sentence and tuple output formats. Models fine-tuned on the sentence format consistently outperform their tuple-based counterparts across most OD tasks on Accuracy (SAMA) (Table~\ref{tab:SAMA Accuracy}). While smaller models show limited improvements, slight enhancements are observed for sentence-format fine-tuning. A similar trend is evident in the train and evaluation loss curves (Supplementary Figure~S1, S2), where sentence-format fine-tuning yields lower losses, though the gap is smaller for MiniCPM-V2.6.

Table~\ref{tab:center_coord_rotation_rmse} demonstrates that models fine-tuned on the sentence format consistently achieve superior numerical approximation performance for center coordinates and rotation angles across all OD tasks. For instance, the RMSE for center coordinates is \textbf{51.64} for the Tuple format compared to \textbf{34.049} for the Sentence format in the MiniCPM-V2.6 model. Additionally, Table~\ref{tab:precission-recall-f1-score} highlights Precision, Recall, and F1 scores, providing another dimension to evaluating the models' effectiveness in attribute prediction and their overall performance. Accuracy (SAMA) evaluates whether attributes are correctly assigned to shapes, while the F1 score measures detection performance irrespective of assignment. Notably, larger models such as MiniCPM-V2.5 and Qwen-2VL 7B show consistent advantages with the sentence format, although MiniCPM-V2.6 exhibits mixed results. For models under 3B parameters, such as Qwen 2B and Paligemma, the tuple format identifies attributes more effectively but struggles with correct alignment, reinforcing trends observed in Section~\ref{Section-4.3}.

The Plant Phenotyping dataset \cite{minervini2016finely} further validates these findings. Table~\ref{tab:model_performance_plant_dataset} reveals that while benefits of the sentence format are minimal when train and test distributions are similar, these advantages become more pronounced under greater distribution shifts. Models fine-tuned on the sentence format achieve lower RMSE values for leaf count prediction in strong OD scenarios.

Finally, we examine the effect of scaled Cross Entropy (CE) Loss for numeric tokens in the LLM tokenizer. Table~\ref{tab:accuracy_scores for scaled loss} demonstrates that scaling the CE loss to values of 2 or 2.5 enhances numerical approximation capabilities, although slight trade-offs in accuracy are observed in Supplementary Table~S2. Furthermore, Supplementary Table~S3 indicates that the Tuple format is less effective than the Sentence format in scaling numeric tokens. These findings suggest that loss scaling plays a critical role in improving numerical reasoning without significantly impacting attribute detection accuracy.

\subsection{Qualitative Results}
\label{Section-4.3}

Supplementary Figures~S3 to S8 compare the sentence and tuple predicted outputs of various Vision-Language Models (VLMs) for a given input image from the validation (evaluation) set of our benchmark. Our analysis reveals that output coherence, defined as the consistency and alignment of predicted attributes with ground truth, improves as model size increases. 

Figures~S3 and S4 highlight the coherent and accurate attribute predictions of larger models such as MiniCPM-V2.6 and MiniCPM-V2.5. These models consistently align predicted attributes with higher comparative accuracy and minimal errors. 

In contrast, smaller models exhibit a noticeable drop in output coherence, as evidenced by Figures~S5, S7, and S8. These figures illustrate common issues, including errors in attribute assignment and misalignment of attributes with shapes. For instance, in Figure~S5 (tuple format), the yellow color is incorrectly assigned to the triangle shape, demonstrating the challenges smaller models face in aligning attributes with the shapes detected.

These results underscore the critical role of model scaling in achieving coherent and accurate predictions, with MiniCPM-V2.6 demonstrating the best performance among the evaluated models.
\section{Discussion}

After extensive experiments, we observe that the output format plays a vital role in fine-tuning Vision-Language Models (VLMs), significantly impacting performance. From Table~\ref{tab:SAMA Accuracy}, we note that larger models fine-tuned in the sentence format outperform those fine-tuned in the tuple format by a wider margin compared to smaller models. Notably, the performance varies across different OD datasets, with models finding datasets containing a higher number of shape interactions particularly challenging. A higher accuracy score (SAMA) indicates that the model not only identifies attributes correctly but also groups them under each detected shape, demonstrating strong spatial and structural understanding.

From Table~\ref{tab:precission-recall-f1-score} we can note that the F1 scores show a similar trend except for MiniCPM-V2.6. Smaller models, such as Paligemma and Qwen-2B (both with fewer than 3B parameters), tend to perform better on this metric when fine-tuned in the tuple format. However, larger models show improved performance across multiple OD datasets when fine-tuned with the sentence format. Notably, MiniCPM-V2.5, based on the popular open-source LLAMA-3 LLM \cite{dubey2024llama}.
This finding becomes particularly notable when we consider the combination of the two metrics discussed above. Accuracy (SAMA) evaluates the model's ability to detect and correctly group attributes for each shape, while the frequency-based F1 score focuses solely on attribute detection without accounting for their structural organization. Our results indicate that smaller models are effective at detecting attributes but struggle to group them correctly, reflecting a lack of strong spatial understanding. Interestingly, for smaller models, the tuple format appears to be more effective for overall attribute detection. In contrast, larger models demonstrate better performance in both attribute detection and structural organization. Notably, fine-tuning larger models using the sentence format further enhances their spatial understanding capabilities.

A key observation from Table~\ref{tab:center_coord_rotation_rmse} highlights the numerical approximation capabilities of VLMs. Sentence format consistently improves numerical predictions for larger VLMs compared to tuple format, with significant improvements observed for MiniCPM-V2.6 and MiniCPM-V2.5. This underscores the importance of output format in enhancing learning and accuracy in numerical tasks.
We extend this experiment to a real-world Plant-Phenotyping dataset, results are shown in Table~\ref{tab:model_performance_plant_dataset} where we observe that sentence format provides a slight improvement in count RMSE. However, under conditions of significant distribution shift, the sentence format achieves better performance and convergence compared to the tuple format.

Supplementary Figures~S1 and S2 illustrate that sentence format for almost all the models leads to a better training and evaluation loss curve during fine-tuning.
Finally, Table~\ref{tab:accuracy_scores for scaled loss} demonstrates that scaling numeric tokens during loss calculation enhances numeric approximation, particularly for the best-performing model, MiniCPM-V2.6.
\section{Conclusion}

This work introduces a novel benchmark to evaluate object attribute detection, spatial understanding, and measurement capabilities of Vision-Language Models (VLMs). Through extensive evaluations of popular open-source VLMs, we demonstrate the significant impact of output formats on fine-tuning performance, with sentence-based formats consistently enhancing both numerical and spatial tasks. Experiments on a real-world plant phenotyping dataset further validate the robustness of our methods under distribution shifts. Additionally, we propose loss scaling strategies to improve numerical approximations and developed effective evaluation methods for compositional outputs, providing valuable tools for advancing spatial reasoning tasks in VLMs. Future work could extend the benchmark to more diverse 2D and 3D shapes and real-world use cases, develop stronger evaluation metrics and fine-tuning strategies, and ultimately enhance the performance of VLMs.



\section{Limitations}

 Dataset limited to five 2D shapes and six colors, with constrained variations in position, occlusion, rotation, and combinations on a 224x224 canvas. Bounding box relaxation may incorrectly classify non-intersecting overlaps, and center-based positions may misrepresent spatial relations for occluded shapes(e.g., left/right or up/down)(see supplementary Section~A2.1). The JV algorithm with edit distance as a cost function may struggle with subtle alignment issues but ensures computational feasibility. Regex patterns perform poorly for unstructured predictions; using LLMs could improve reliability, but with high compute costs. Precision, Recall, and F1 scores may overestimate performance by not evaluating attribute-shape assignments; integrating them with SAMA Accuracy addresses both detection and assignment accuracy.
{\small
\bibliographystyle{ieee}
\bibliography{mpbib}
}

\clearpage
\appendix      
\section{Appendix}
\label{sec:appendix}
\makeatletter
  \renewcommand\thefigure{S\arabic{figure}}
  \renewcommand\thetable{S\arabic{table}}
  \setcounter{figure}{0}
  \setcounter{table}{0}
\makeatother

\subsection{Ablation Studies}

Table~\ref{tab:ablation_study_for_FT} shows that fine-tuning only the vision head or the language model (LLM) component of the VLM results in a decrease in performance. Specifically, fine-tuning only the vision head leads to a significant performance drop, whereas fine-tuning only the language component causes a relatively minor decline. These findings indicate that fine-tuning both components jointly is beneficial for our task.

Table~\ref{tab:accuracy-center-rotation-rmse-scaled-Tuple-vs-sentence} demonstrates that loss scaling is more effective for the Sentence Format compared to the Tuple Format. The efficacy of loss scaling is also contingent upon the specific out-of-domain (OD) scenarios under examination, with notable advantages observed in cases such as Occlusion, Rotation, and Spatial variations. It is important to note that while scaling may induce a slight reduction in Accuracy (SAMA), the scale parameter can be fine-tuned to achieve an optimal balance between these metrics, as illustrated in Table~\ref{tab:accuracy-center-rotation-rmse-scaled}.

\begin{table*}[h]
   \centering
   \small
   \begin{tabular}{lcccccc}
       \toprule
       \textbf{Model} & \textbf{OD Comp.} & \textbf{OD Occl.} & \textbf{OD Rot.} & \textbf{OD Spatial} & \textbf{Test Set} & \textbf{OD Size} \\
       \midrule
       \multicolumn{7}{c}{\textit{SAMA Accuracy $\uparrow$}} \\
       \midrule
       MiniCPM-V2.6 (Sentence) & \textbf{0.532} & \textbf{0.694} & \textbf{0.679} & \textbf{0.531} & \textbf{0.694} & 0.608 \\
       MiniCPM-V2.6 (LLM only) & 0.522 & 0.673 & 0.657 & 0.529 & 0.661 & \textbf{0.614} \\
       MiniCPM-V2.6 (VPM Only) & 0.367 & 0.565 & 0.587 & 0.356 & 0.583 & 0.068 \\
       \midrule
       \multicolumn{7}{c}{\textit{Center RMSE $\downarrow$}} \\
       \midrule
       MiniCPM-V2.6 (Sentence) & \textbf{33.54} & \textbf{23.49} & \textbf{22.66} & \textbf{35.50} & \textbf{20.03} & \textbf{79.10} \\
       MiniCPM-V2.6 (LLM only) & 37.71 & 28.81 & 27.71 & 39.01 & 24.03 & 80.55 \\
       MiniCPM-V2.6 (VPM Only) & 75.43 & 64.32 & 62.36 & 76.14 & 64.96 & 167.66 \\
       \midrule
       \multicolumn{7}{c}{\textit{Rotation RMSE $\downarrow$}} \\
       \midrule
       MiniCPM-V2.6 (Sentence) & 16.80 & \textbf{3.92} & \textbf{22.49} & \textbf{4.65} & \textbf{4.53} & \textbf{6.45} \\
       MiniCPM-V2.6 (LLM only) & \textbf{16.20} & 5.25 & 23.16 & 5.55 & 5.10 & 6.90 \\
       MiniCPM-V2.6 (VPM Only) & 19.88 & 9.98 & 26.11 & 6.38 & 8.78 & 8.63 \\
       \bottomrule
   \end{tabular}

   \caption{Performance comparison of MiniCPM-V2.6 variants across different metrics and test sets. For SAMA Accuracy higher is better (↑), while for Center and Rotation RMSE lower is better (↓).}
   \label{tab:ablation_study_for_FT}
\end{table*}

\subsection{Implementation Details}
\subsubsection{Benchmark Dataset:}

The training dataset comprises images containing up to five distinct shapes: Circle, Rectangle, Ellipse, Triangle, and Square. These shapes are specifically selected to facilitate the interpretation of rotational transformations. Occlusions are incorporated, with a maximum of 2 to 3 overlapping shapes per image. Each image contains between 2 to 4 shapes, with each shape randomly assigned one of six colors: Orange, Red, Blue, Green, Yellow, or Magenta. Rotational transformations of 0$^\circ$, 15$^\circ$, and 30$^\circ$ are applied to the shapes without introducing rotational symmetries that could result in equivalent orientations. For instance, when a square is rotated by 15$^\circ$, it is explicitly labeled as a 15$^\circ$ rotation and is not treated as equivalent to rotations such as 105$^\circ$, 195$^\circ$, or 285$^\circ$, which would appear identical due to the symmetry of the square. Thus, each rotation is uniquely assigned to its specific angle, ensuring that a shape rotated by 15$^\circ$ is always recognized as a 15$^\circ$ rotation and not as an equivalent symmetric angle. Circles, due to their symmetry, are always assigned a rotation of 0$^\circ$.

\begin{figure*}[p]
    \centering
    \includegraphics[width=\linewidth,height=1\textheight,keepaspectratio]{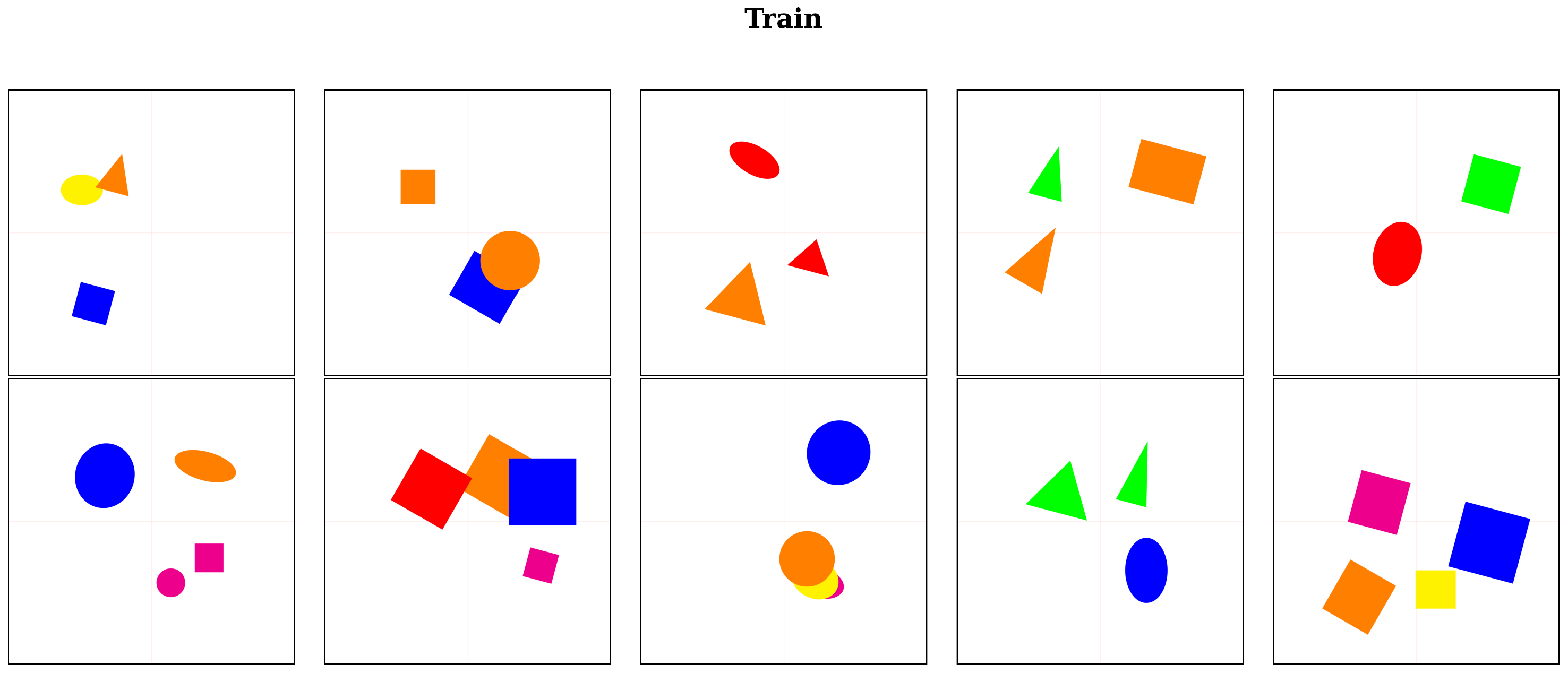}
    \caption{Examples of training set images from the benchmark, illustrating shape configurations, occlusions, and attribute variations used during model training.}
    \label{fig:example_train}
\end{figure*}

In this context, "occlusion" refers to the maximum number of shapes allowed to overlap in a single configuration. For example, an occlusion limit of 3 implies that up to three shapes can overlap in a single image, but no more than three shapes will participate in any one overlapping arrangement. Ground‑truth occlusion is determined by checking if the shape's bounding boxes overlap. However, this can incorrectly indicate occlusion for shapes like circles whose bounding boxes overlap despite the actual shapes not touching. To address this, we apply bounding‑box relaxation: we ignore slight bounding‑box overlaps, thus reducing false positives in occlusion detection. A "configuration" denotes the set of parameters that define each shape and control its arrangement within the image. The target images are rendered using TikZ based on these configurations, where each configuration specifies the attributes required to generate the corresponding TikZ image. Examples are provided in Figure~\ref{fig:example_train}
The evaluation dataset shares the same configuration parameters as the training dataset but contains unique and exclusive examples. To ensure there are no duplicates, the uniqueness of each image and its corresponding configuration is verified using the MD5 hashing algorithm. While the hashing algorithm helps prevent duplicate configurations, even a minor change in an otherwise identical setting will still hash to a different value. Though this is a rare limitation, in practice it enables us to generate sufficiently diverse samples (see Figure~\ref{fig:example_train}).

\paragraph{OD Scenarios}

\begin{figure*}[p]
    \centering
    \includegraphics[width=\linewidth,height=1\textheight,keepaspectratio]{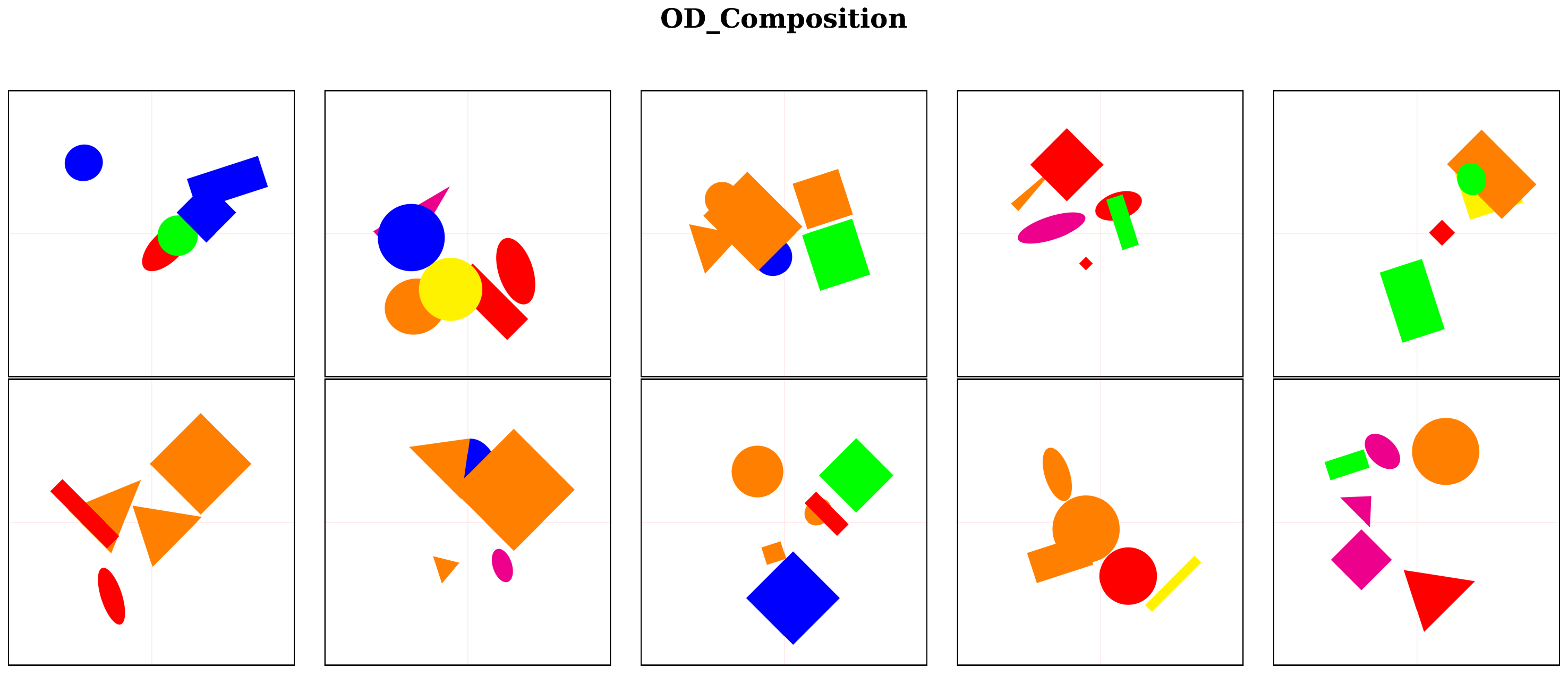}
    \caption{Benchmark: OD Composition Set Examples}
    \label{fig:example_od_composition}
\end{figure*}

\begin{figure*}[p]
    \centering
    \includegraphics[width=\linewidth,height=1\textheight,keepaspectratio]{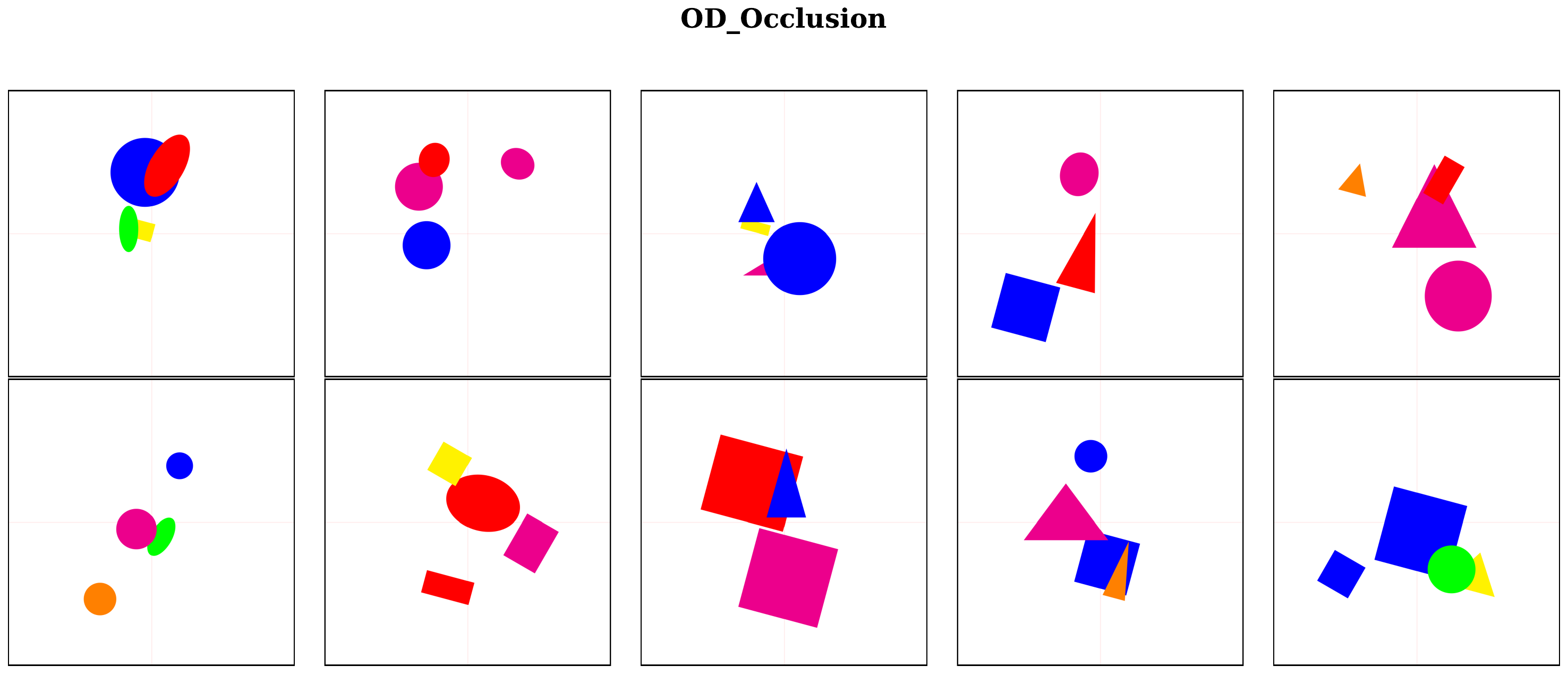}
    \caption{Benchmark: OD Occlusion Set Examples}
    \label{fig:example_od_occlusion}
\end{figure*}

\begin{figure*}[p]
    \centering
    \includegraphics[width=\linewidth,height=1\textheight,keepaspectratio]{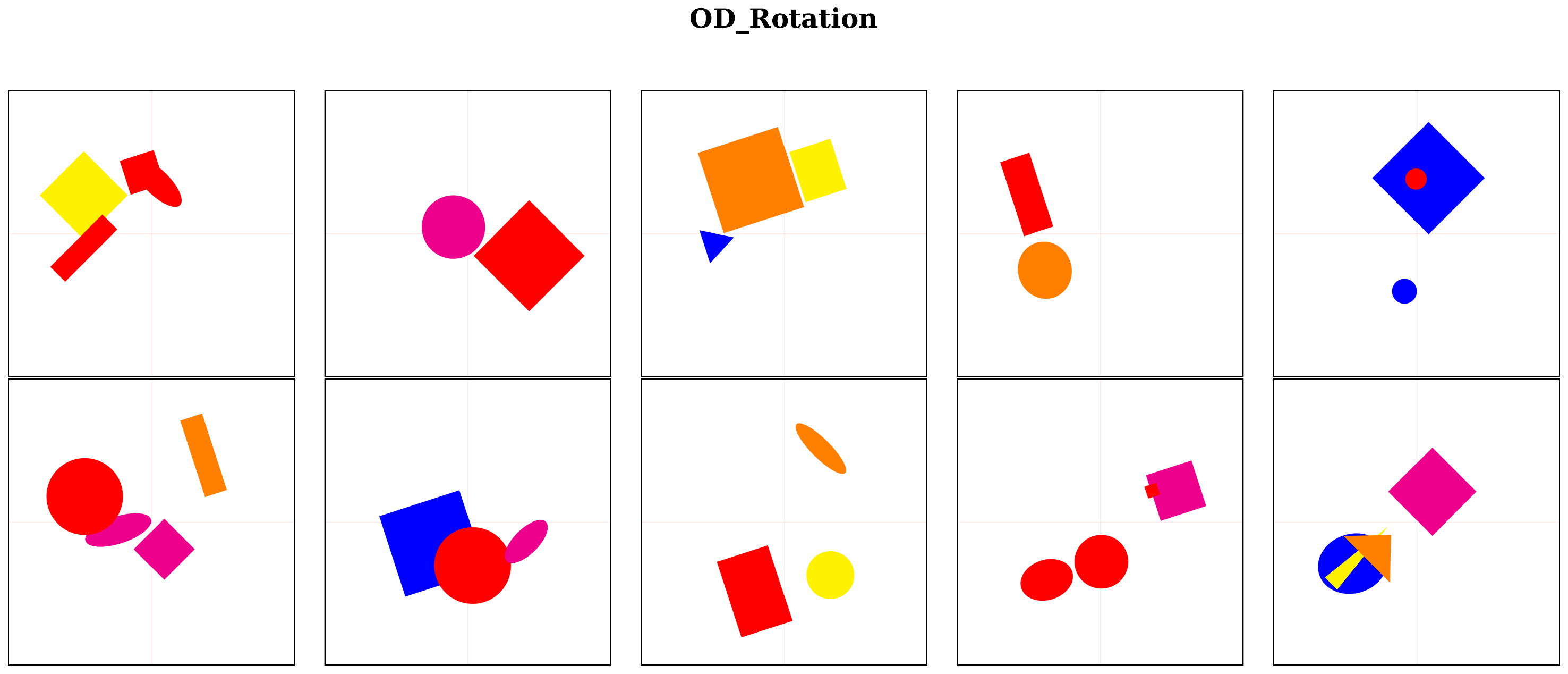}
    \caption{Benchmark: OD Rotation Set Examples}
    \label{fig:example_od_rotaion}
\end{figure*}

\begin{figure*}[p]
    \centering
    \includegraphics[width=\linewidth,height=1\textheight,keepaspectratio]{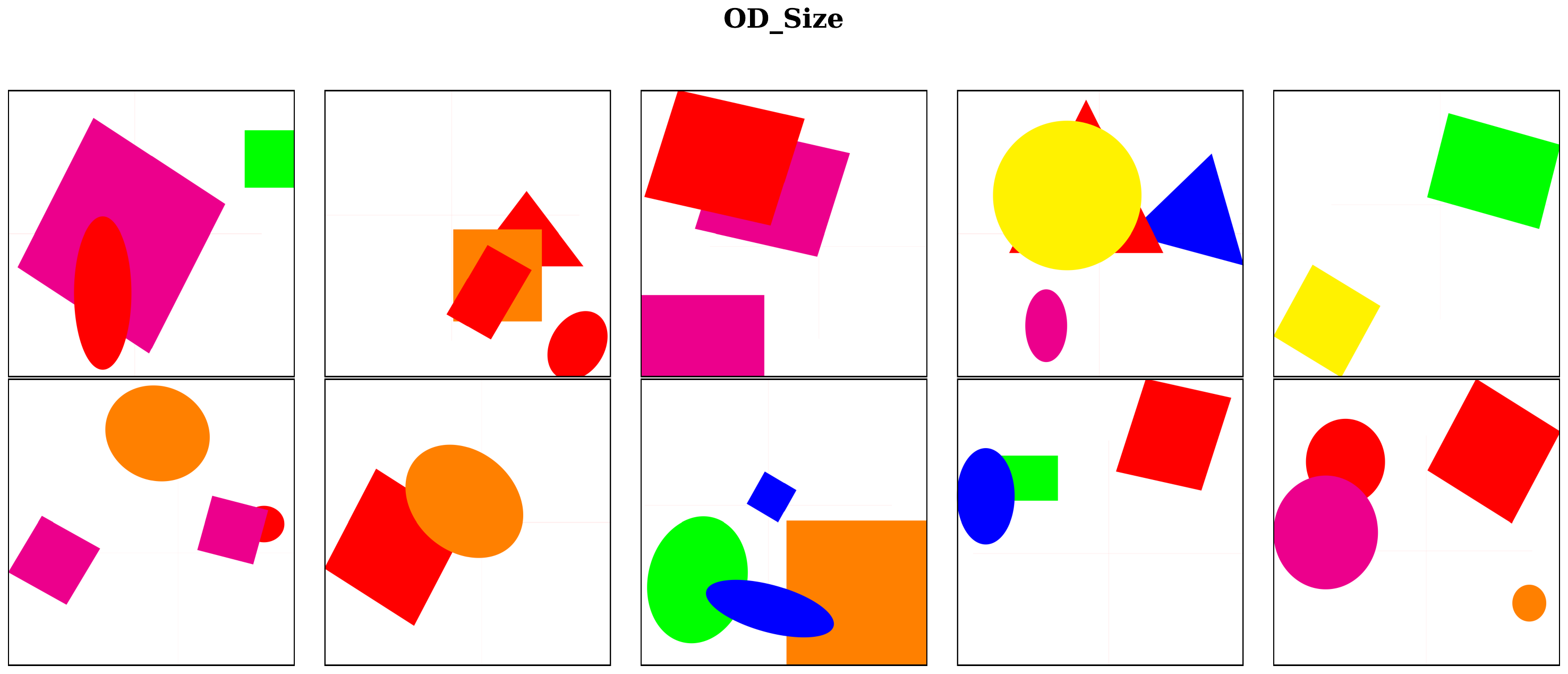}
    \caption{Benchmark: OD Size Set Examples}
    \label{fig:example_od_size}
\end{figure*}

\begin{figure*}[p]
    \centering
    \includegraphics[width=\linewidth,height=1\textheight,keepaspectratio]{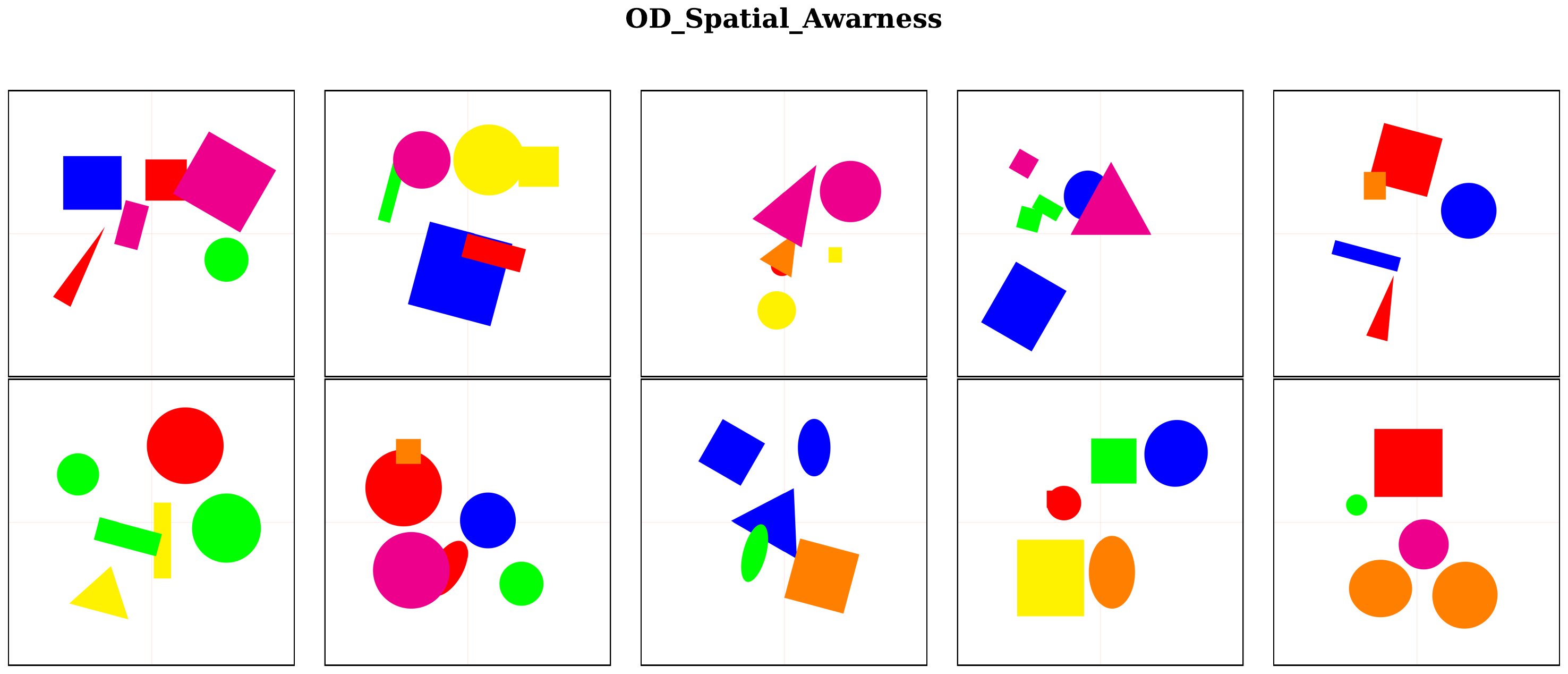}
    \caption{Benchmark: OD Spatial Awareness Set Examples}
    \label{fig:example_od_spatial_awarness}
\end{figure*}

We define five distinct Out-of-Domain (OD) scenarios to evaluate the generalization capabilities of the models:

\begin{enumerate}
    \item \textbf{OD Compositional}: This scenario relaxes the constraints on the number of shapes and the limit on overlapping shapes from the training set. Specifically, the shape limit is increased to 5–6 shapes per image, and the occlusion limit is also raised to 5–6 shapes. This configuration challenges the model to handle a greater number of shapes per image and to recognize attributes under more complex overlap scenarios not encountered during training. Examples are illustrated in Figure~\ref{fig:example_od_composition}.
    
    \item \textbf{OD Spatial Awareness}: In this scenario, the number of shapes per image is similarly increased to 5–6, introducing new spatial arrangements. This setup tests the model's ability to detect and attribute more shapes in configurations that were not present in the training data. The scenario evaluates the model’s capacity to generalize to novel spatial distributions. Examples are provided in Figure~\ref{fig:example_od_spatial_awarness}.
    
    \item \textbf{OD Occlusion}: While maintaining the same basic configuration as the training set, this scenario relaxes the occlusion limit, allowing for 4–5 shapes to overlap simultaneously. This scenario assesses the model’s ability to accurately detect and attribute shapes under previously unseen levels of occlusion. Examples are depicted in Figure~\ref{fig:example_od_occlusion}.
    
    \item \textbf{OD Rotation}: This scenario introduces additional rotation angles (45$^\circ$ and 72$^\circ$) for the shapes, presenting a challenge for the models to identify shapes in new orientations and accurately predict their angles. The ability to generalize to these out-of-domain rotations is evaluated in this scenario. Examples are shown in Figure~\ref{fig:example_od_rotaion}.
    
    \item \textbf{OD Size}: This scenario scales all shapes by a factor of 2, resulting in shapes that are twice their original size. It challenges the models to accurately locate the shapes and predict their attributes and coordinates under this new size configuration. Examples are provided in Figure~\ref{fig:example_od_size}.
\end{enumerate}

We generate the training dataset with two distinct output formats: sentence format and tuple format.

In the sentence format, the target ground truth is generated using a predefined template, where variables are populated with the relevant attributes from the configuration of the corresponding image. The template is as follows:

\texttt{
A \{Color[i]\} \{shapes[i]\} is located in the \{quadrant[i]\} quadrant, \\
centred at coordinates \{Center\_coordinates[i]\}, with relative \\
positions described as \{Relative\_Position[i]\}, rotated by \{Rotation[i]\} \\
degrees, and is \{'occluded' if Occlusion[i] == 'Yes' else 'not occluded'\}.
}

In the tuple format, the attributes are represented in a structured format similar to JSON, where variables are also populated from the configuration. The format is as follows:

\texttt{
({shapes[i]}, \\
quadrant={quadrant[i]}, \\
center\_coordinates={Center\_coordinates[i]}, \\
relative\_position={Relative\_Position[i]}, \\
rotation={Rotation[i]}, \\
occlusion={Occlusion[i]}, \\
color={Color[i]})
}

Both formats are designed to evaluate different Vision-Language Models (VLMs). The sentence format aligns with natural language, while the tuple format provides a structured representation, enabling us to study their impacts.

\begin{figure*}[p]
    \centering
    \includegraphics[width=\linewidth,height=1\textheight,keepaspectratio]{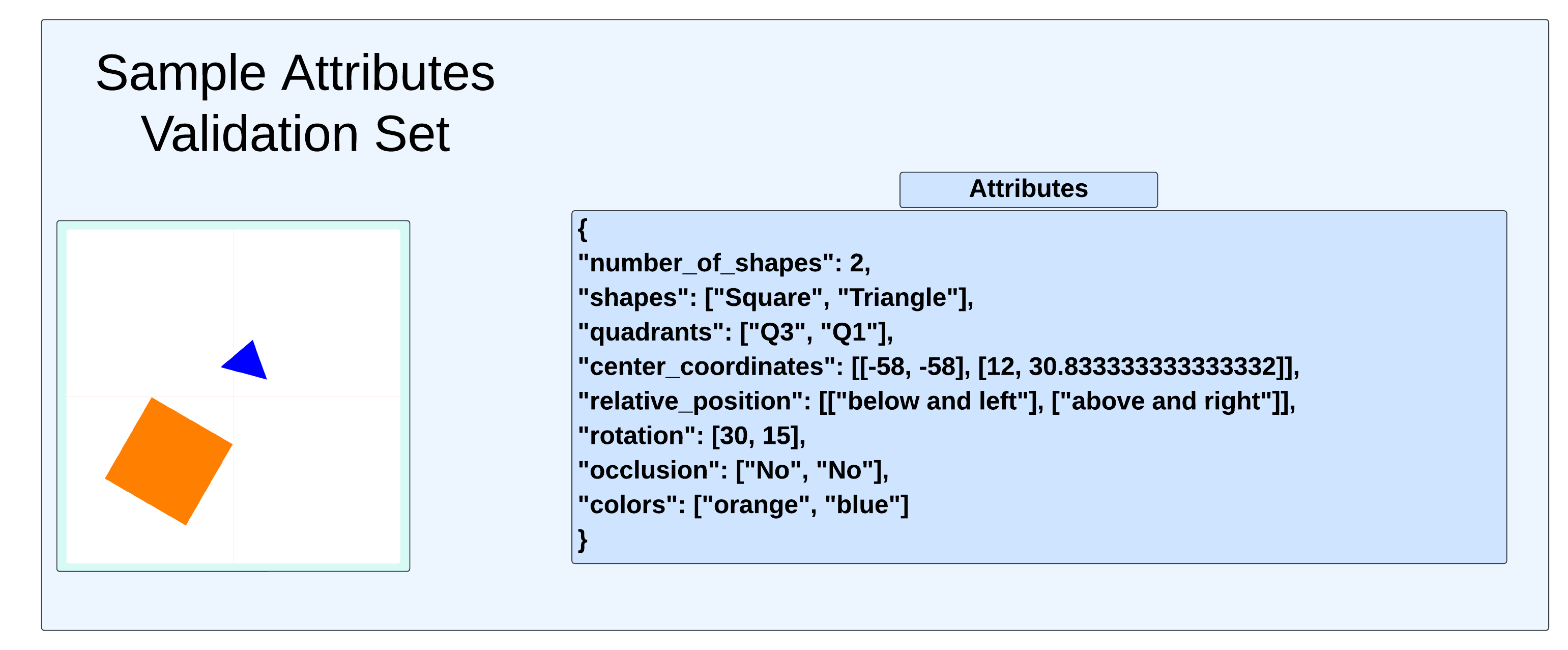}
    \caption{Benchmark: Train Set Examples}
    \label{fig:Dataset_example}
\end{figure*}

Figure~\ref{fig:Dataset_example} provides an illustrative example of the attribute configuration used for both formats.

\subsubsection{Applied Evaluation: Plant Phenotyping Dataset}

\begin{figure*}[p]
    \centering
    \includegraphics[width=\linewidth,height=1\textheight,keepaspectratio]{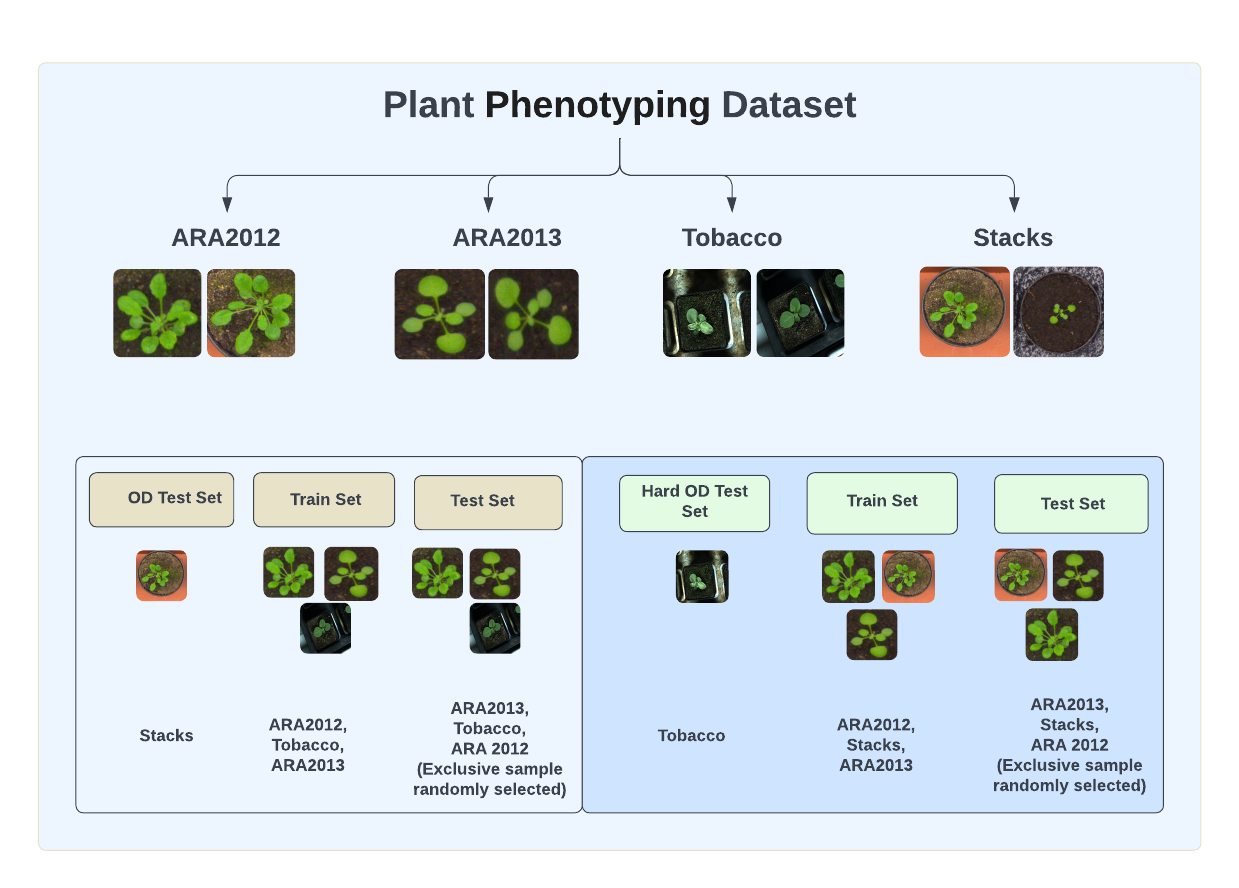}
    \caption{Visualization of the process used to create OD test sets for the Plant Phenotyping Dataset, demonstrating subsets and domain shifts in image styles.}
    \label{fig:Plant_phenotyping_dataset}
\end{figure*}

We evaluate our approach using the Plant Phenotyping Dataset, which comprises four subsets: ARA2012, ARA2013, Tobacco, and Stacks (the latter containing both ARA2012 and ARA2013). Various combinations of these subsets are utilized to construct distinct training sets and corresponding out-of-domain (OD) test sets. Specifically, we design an OD test set with a smaller domain shift as well as a "hard" OD test set with a significant domain shift, characterized by differences in image styles. This process is illustrated in Figure~\ref{fig:Plant_phenotyping_dataset}.

\begin{figure*}[p]
    \centering
    \includegraphics[width=\linewidth,height=1\textheight,keepaspectratio]{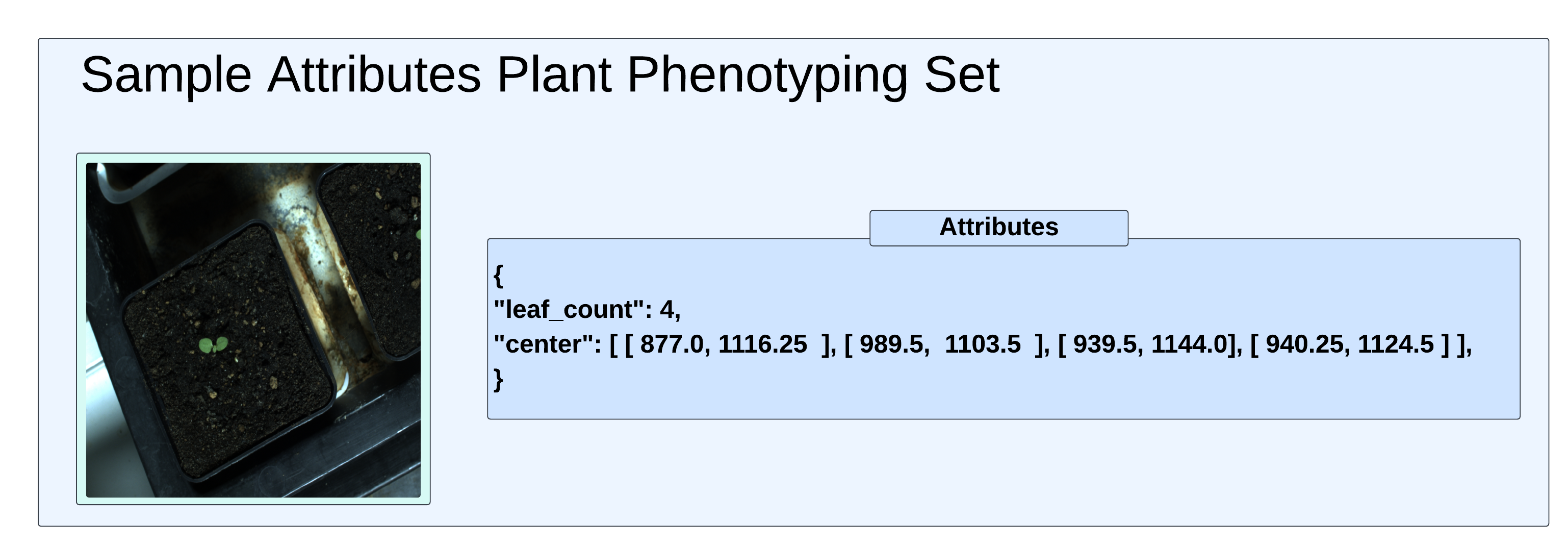}
    \caption{Illustration of attributes in the Plant Phenotyping Dataset, including leaf counts, Center Coordinates, and derived features for evaluation tasks.}
    \label{fig:Plant_phenotyping_dataset_attributes}
\end{figure*}

The model is fine-tuned to predict the number of leaves in a given image, as depicted in Figure~\ref{fig:Plant_phenotyping_dataset_attributes}. This evaluation assesses the model's generalization capability in real-world OD scenarios, testing its performance under both small and large distribution shifts. Furthermore, we analyze the impact of the two output formats (sentence and tuple) on model performance. Our findings indicate that the sentence format achieves superior results compared to the tuple format under conditions of large distribution shift, as demonstrated in Table~3. 
A discrepancy in center coordinate RMSE, as noted in Table~\ref{tab:model_performance_plant_dataset_table_2_supp}, is likely attributed to the use of linear matching. In this approach, RMSE is calculated based on the closest predictions to ground truth pairs, even when some leaves remain undetected. This behavior highlights the need for further investigation to refine and improve evaluation methodologies.

\subsection{Open‐Source Model Licensing }
All pretrained vision–language models used in this study (MiniCPM-V2.5, MiniCPM-V2.6, Qwen2-VL-2B, Qwen2-VL-7B, Phi-3.5-vision-instruct, and PaliGemma-3b-pt-224) are publicly released under the Apache License 2.0.

\subsection{Use of AI Assistants }
We used GitHub Copilot and ChatGPT to help refine our experimental code, assist with debugging, and polish draft text; all AI‐generated suggestions were reviewed and edited by the authors to ensure correctness and clarity.

\subsection{Additional Results}

\begin{table*}[t]
    \centering
    \small
    \begin{tabular}{l*{9}{c}}
        \toprule
        & \multicolumn{3}{c}{\textbf{OD Comp.}} & \multicolumn{3}{c}{\textbf{OD Occl.}} & \multicolumn{3}{c}{\textbf{OD Rot.}} \\
        \cmidrule(lr){2-4} \cmidrule(lr){5-7} \cmidrule(lr){8-10}
        \textbf{Model} & SAMA↑ & C↓ & \cellcolor{gray!20}R↓ & SAMA↑ & C↓ & \cellcolor{gray!20}R↓ & SAMA↑ & C↓ & \cellcolor{gray!20}R↓ \\
        \midrule
        MiniCPM-V2.6 & 0.504 & 51.64 & \cellcolor{gray!20}34.8 & 0.654 & 32.27 & \cellcolor{gray!20}15.57 & 0.649 & 33.71 & \cellcolor{gray!20}36.09 \\
        MiniCPM-V2.6 [SF] & \textbf{0.547} & 34.05 & \cellcolor{gray!20} \textbf{16.12} &  \textbf{0.699} & 24.22 & \cellcolor{gray!20}4.25 & 0.688 & 23.29 & \cellcolor{gray!20}22.06 \\
        \midrule
        MiniCPM-V2.6 SW scale 1.5 & 0.527 & 35.19 & \cellcolor{gray!20}16.35 & 0.690 & 22.45 & \cellcolor{gray!20}4.35 & \textbf{0.690} &  \textbf{21.85} & \cellcolor{gray!20}21.43 \\
        MiniCPM-V2.6 SW scale 2 & 0.534 & \textbf{31.81} & \cellcolor{gray!20}16.65 & 0.675 & \textbf{21.53} & \cellcolor{gray!20}4.20 & 0.663 & 22.94 & \cellcolor{gray!20}21.66 \\
        MiniCPM-V2.6 SW scale 2.5 & 0.523 & 34.66 & \cellcolor{gray!20}16.73 & 0.696 & 21.67 & \cellcolor{gray!20}4.80 & 0.681 & 22.46 & \cellcolor{gray!20}22.11 \\
        MiniCPM-V2.6 SW scale 3.5 & 0.514 & 36.53 & \cellcolor{gray!20}17.03 & 0.678 & 22.52 & \cellcolor{gray!20} \textbf{4.05} & 0.664 & 22.02 & \cellcolor{gray!20}\textbf{21.43} \\
        \midrule
        & \multicolumn{3}{c}{\textbf{OD Spatial}} & \multicolumn{3}{c}{\textbf{Test Set}} & \multicolumn{3}{c}{\textbf{OD Size}} \\
        \cmidrule(lr){2-4} \cmidrule(lr){5-7} \cmidrule(lr){8-10}
        \textbf{Model} & SAMA↑ & C↓ & \cellcolor{gray!20}R↓ & SAMA↑ & C↓ & \cellcolor{gray!20}R↓ & SAMA↑ & C↓ & \cellcolor{gray!20}R↓ \\
        \midrule
        MiniCPM-V2.6 & 0.517 & 55.37 & \cellcolor{gray!20}58.42 & 0.654 & 32.63 & \cellcolor{gray!20}56.27 & \textbf{0.605} & 98.10 & \cellcolor{gray!20}6.50 \\
        MiniCPM-V2.6 [SF] &  \textbf{0.541} & 34.13 & \cellcolor{gray!20}4.64 & 0.688 & 19.77 & \cellcolor{gray!20}4.57 & 0.600 & 80.66 & \cellcolor{gray!20}6.05 \\
        \midrule
        MiniCPM-V2.6 SW scale 1.5 & 0.537 & \textbf{31.56} & \cellcolor{gray!20}4.80 & \textbf{0.691} & 18.89 & \cellcolor{gray!20}4.74 & 0.593 & 80.77 & \cellcolor{gray!20}6.08 \\
        MiniCPM-V2.6 SW scale 2 & 0.536 & 32.44 & \cellcolor{gray!20}5.10 & 0.679 & 18.45 & \cellcolor{gray!20}4.28 & 0.581 & 82.39 & \cellcolor{gray!20}5.63 \\
        MiniCPM-V2.6 SW scale 2.5 & 0.538 & 32.77 & \cellcolor{gray!20}5.03 & 0.684 & 18.36 & \cellcolor{gray!20}4.68 & 0.596 & 82.37 & \cellcolor{gray!20}5.55 \\
        MiniCPM-V2.6 SW scale 3.5 & 0.525 & 36.02 & \cellcolor{gray!20}\textbf{4.64} & 0.675 & \textbf{18.21} & \cellcolor{gray!20}\textbf{4.28} & 0.583 & \textbf{79.38} & \cellcolor{gray!20}\textbf{5.55} \\
        \bottomrule
    \end{tabular}
    \caption{Performance comparison across different model variations, showing Structured Attribute Matching Accuracy (SAMA), Center RMSE (C), and Rotation RMSE (R). ↑ indicates higher values are better, ↓ indicates lower values are better. SW stands for Sentence Weighted and SF stands for Sentence Format. Base model is Tuple Format.}
    \label{tab:accuracy-center-rotation-rmse-scaled}
\end{table*}

\begin{table*}[t]
    \centering
    \small
    \begin{tabular}{l*{9}{c}}
        \toprule
        & \multicolumn{3}{c}{\textbf{OD Comp.}} & \multicolumn{3}{c}{\textbf{OD Occl.}} & \multicolumn{3}{c}{\textbf{OD Rot.}} \\
        \cmidrule(lr){2-4} \cmidrule(lr){5-7} \cmidrule(lr){8-10}
        \textbf{Model} & SAMA↑ & C↓ & R↓ & SAMA↑ & C↓ & R↓ & SAMA↑ & C↓ & R↓ \\
        \midrule
        MiniCPM-V2.6 [SF] & \textbf{0.547} & \textbf{34.05} & \textbf{16.12} & \textbf{0.699} & 24.22 & \textbf{4.25} & 0.688 & 23.29 & 22.06 \\
        MiniCPM-V2.6 SW & 0.527 & 35.19 & 16.35 & 0.690 & \textbf{22.45} & 4.35 & \textbf{0.690} & \textbf{21.85} & \textbf{21.43} \\
        \midrule
        MiniCPM-V2.6 & 0.504 & \textbf{51.64} & 34.81 & 0.654 & 32.27 & 5.58 & \textbf{0.649} & \textbf{33.71} & 36.10 \\
        MiniCPM-V2.6 TW & \textbf{0.505} & 55.66 & \textbf{34.79} & \textbf{0.654} & \textbf{32.13} & \textbf{5.48} & 0.641 & 34.44 & \textbf{35.74} \\
        \bottomrule
        \addlinespace[1em]
        & \multicolumn{3}{c}{\textbf{OD Spatial}} & \multicolumn{3}{c}{\textbf{Test Set}} & \multicolumn{3}{c}{\textbf{OD Size}} \\
        \cmidrule(lr){2-4} \cmidrule(lr){5-7} \cmidrule(lr){8-10}
        \textbf{Model} & SAMA↑ & C↓ & R↓ & SAMA↑ & C↓ & R↓ & SAMA↑ & C↓ & R↓ \\
        \midrule
        MiniCPM-V2.6 [SF] & \textbf{0.541} & 34.13 & \textbf{4.60} & 0.688 & 19.77 & \textbf{4.57} & \textbf{0.600} & \textbf{80.66} & \textbf{6.05} \\
        MiniCPM-V2.6 SW & 0.537 & \textbf{31.56} & 4.80 & \textbf{0.691} & \textbf{18.89} & 4.74 & 0.593 & 80.77 & 6.08 \\
        \midrule
        MiniCPM-V2.6 & 0.517 & \textbf{55.37} & 8.425 & \textbf{0.654} & 32.63 & 6.27 & \textbf{0.605} & 98.10 & 6.50 \\
        MiniCPM-V2.6 TW & \textbf{0.521} & 55.57 & \textbf{8.17} & 0.646 & \textbf{32.07} & \textbf{5.85} & 0.583 & \textbf{96.55} & \textbf{6.38} \\
        \bottomrule
    \end{tabular}
    \caption{Performance evaluation across OD scenarios using Structured Attribute Matching Accuracy (SAMA), Center RMSE (C), and Rotation RMSE (R). SAMA (↑) measures attribute matching accuracy, while C (↓) and R (↓) quantify spatial and rotational errors. SF denotes Sentence Format, SW indicates Sentence Weighted, and TW  indicates Tuple Weighted. Arrows indicate better performance direction. The scale value used is 1.5 }
    \label{tab:accuracy-center-rotation-rmse-scaled-Tuple-vs-sentence}
\end{table*}


\begin{table*}[h]
    \centering
    \small
    \begin{tabular}{l|cc|cc|cc|cc}
        \toprule
        & \multicolumn{2}{c|}{\textbf{Test Set}} & \multicolumn{2}{c|}{\textbf{Train Set}} & \multicolumn{2}{c|}{\textbf{OD Test Set}} & \multicolumn{2}{c}{\textbf{Strong OD Test Set}} \\
        \textbf{Model} & \textbf{Count↓} & \textbf{Center↓} & \textbf{Count↓} & \textbf{Center↓} & \textbf{Count↓} & \textbf{Center↓} & \textbf{Count↓} & \textbf{Center↓} \\
        \midrule
        Plant\_sentence & \textbf{1.19} & 152.31 & \textbf{1.78} & 238.48 & \textbf{1.59} & \textbf{87.15} & \textbf{3.01} & 1240.96 \\
        Plant\_Tuple & 1.31 & \textbf{121.78} & 1.99 & \textbf{219.00} & 1.65 & 93.32 & 6.28 & \textbf{1233.40} \\
        \bottomrule
    \end{tabular}
    
    \caption{Performance comparison between sentence and tuple formats on the Plant Phenotyping dataset. The gap widens in favour of sentence format as out-of-distribution difficulty increases. ↓: lower is better. Count: Count RMSE, Center: Center RMSE.}
    \label{tab:model_performance_plant_dataset_table_2_supp}
\end{table*}

\clearpage  
\begin{figure*}[p]  
    \centering
    \includegraphics[width=\linewidth,height=\textheight,keepaspectratio]{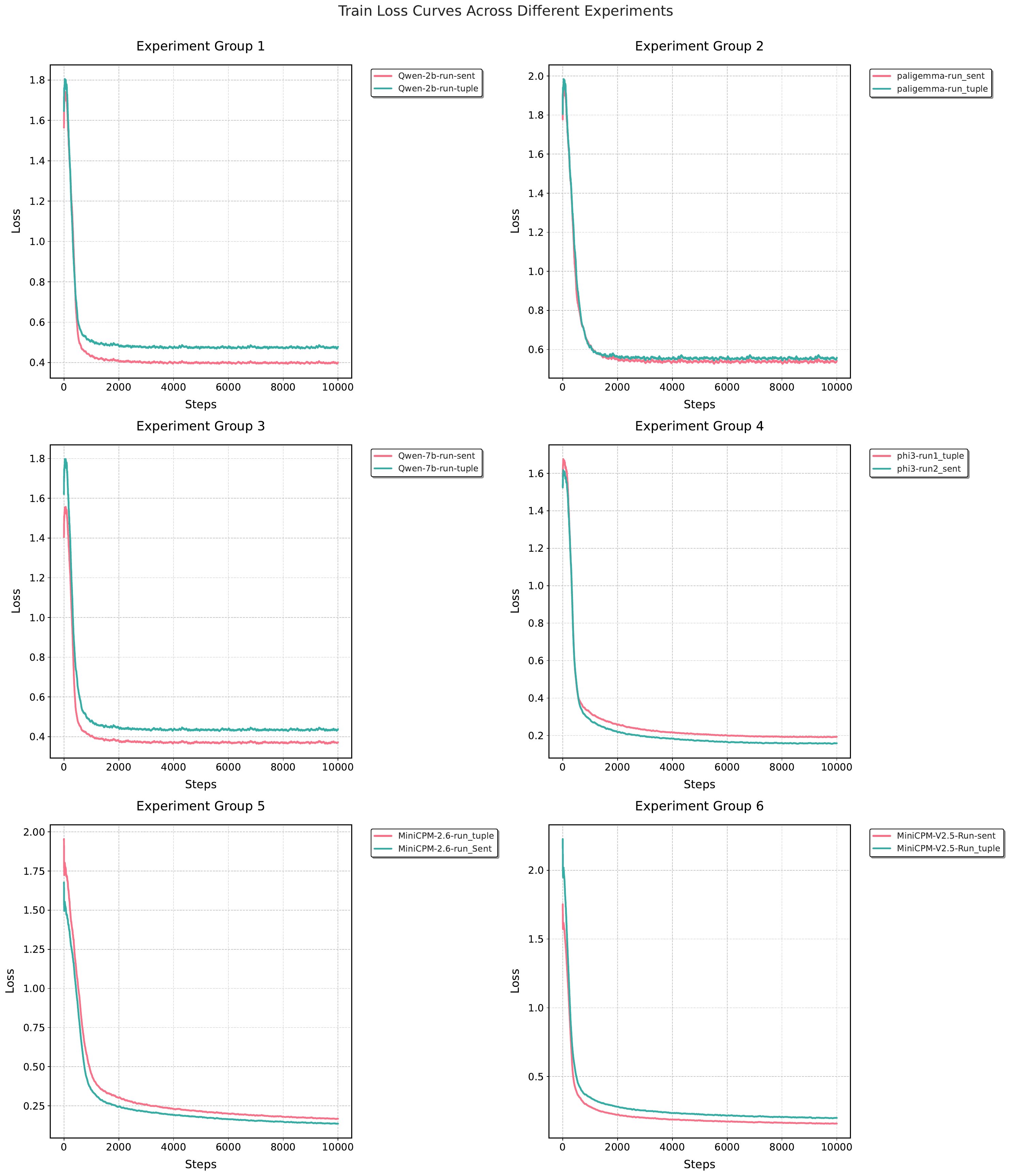}
    \caption{Training loss curves across different experimental settings. Each subplot compares two variants of the same experimental setup, showing the convergence behavior during training.}
    \label{fig:train_curves}
\end{figure*}

\clearpage  

\clearpage  
\begin{figure*}[p]
    \centering
    \includegraphics[width=\linewidth,height=\textheight,keepaspectratio]{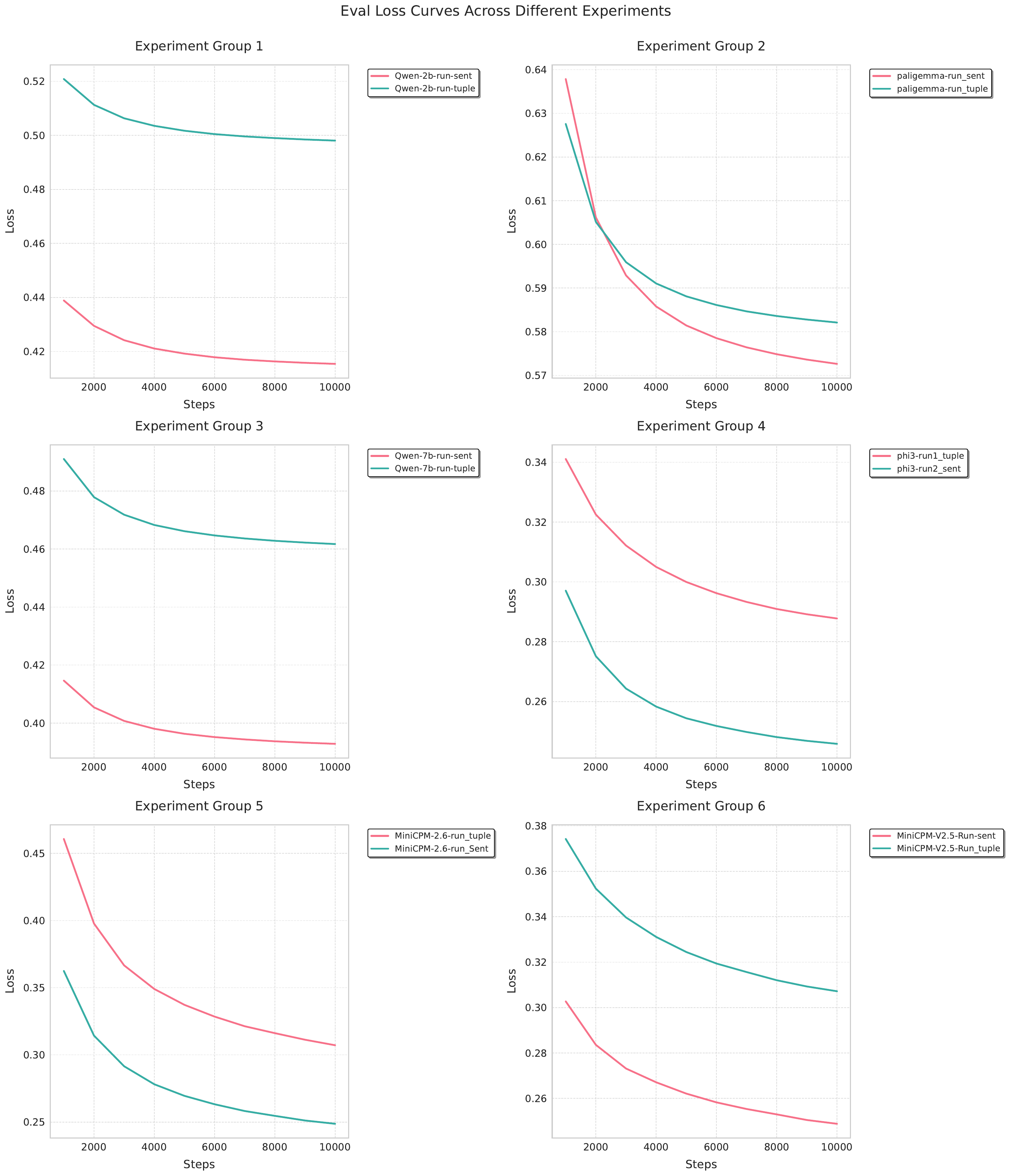}
    \caption{Evaluation loss curves across different experimental settings. Each subplot compares two variants of the same experimental setup, demonstrating the generalization behavior of different model configurations.}
    \label{fig:eval_curves}
\end{figure*}



\begin{figure*}[p]
    \centering
    \includegraphics[width=\linewidth,height=1\textheight,keepaspectratio]{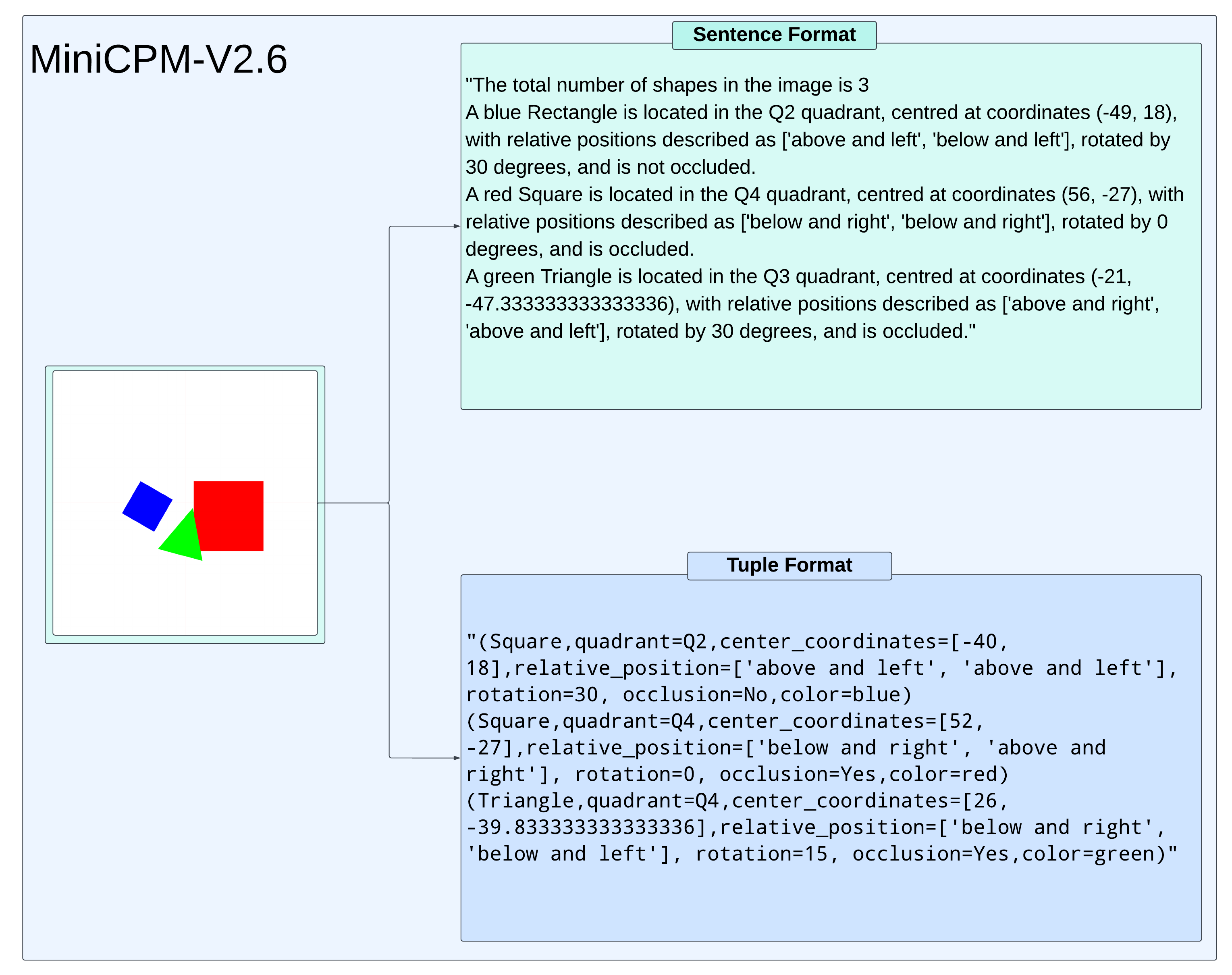}
    \caption{Sentence Vs Tuple Output Comparison for MiniCPM-V2.6 Models for Validation Dataset}
    \label{fig:sentence_vs_tuple_figure_all_models}
\end{figure*}


\begin{figure*}[p]
    \centering
    \includegraphics[width=\linewidth,height=1\textheight,keepaspectratio]{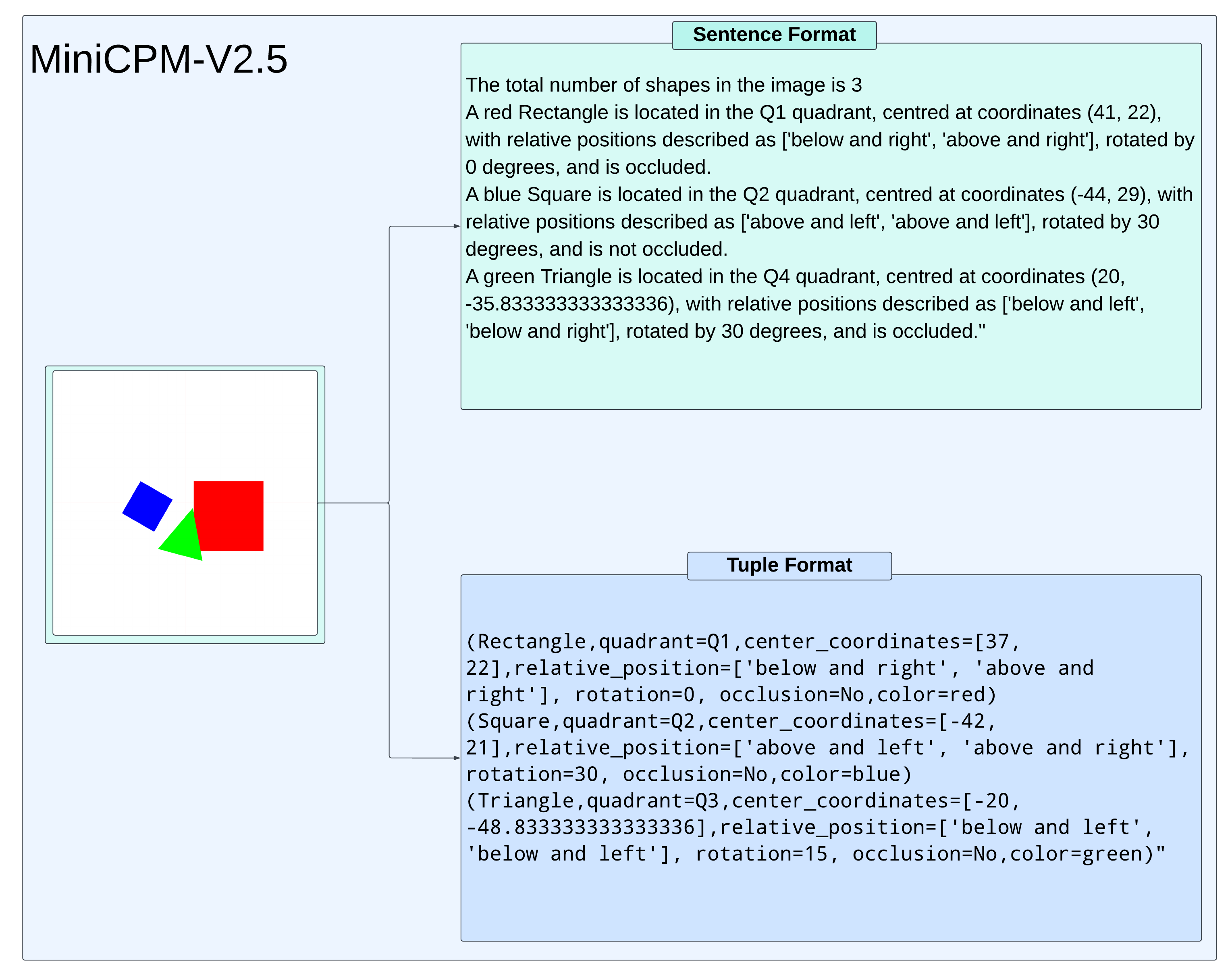}
    \caption{Sentence Vs Tuple Output Comparison for MiniCPM-V2.5 Models for Validation Dataset}
    \label{fig:sentence_vs_tuple_figure_all_models}
\end{figure*}


\begin{figure*}[p]
    \centering
    \includegraphics[width=\linewidth,height=1\textheight,keepaspectratio]{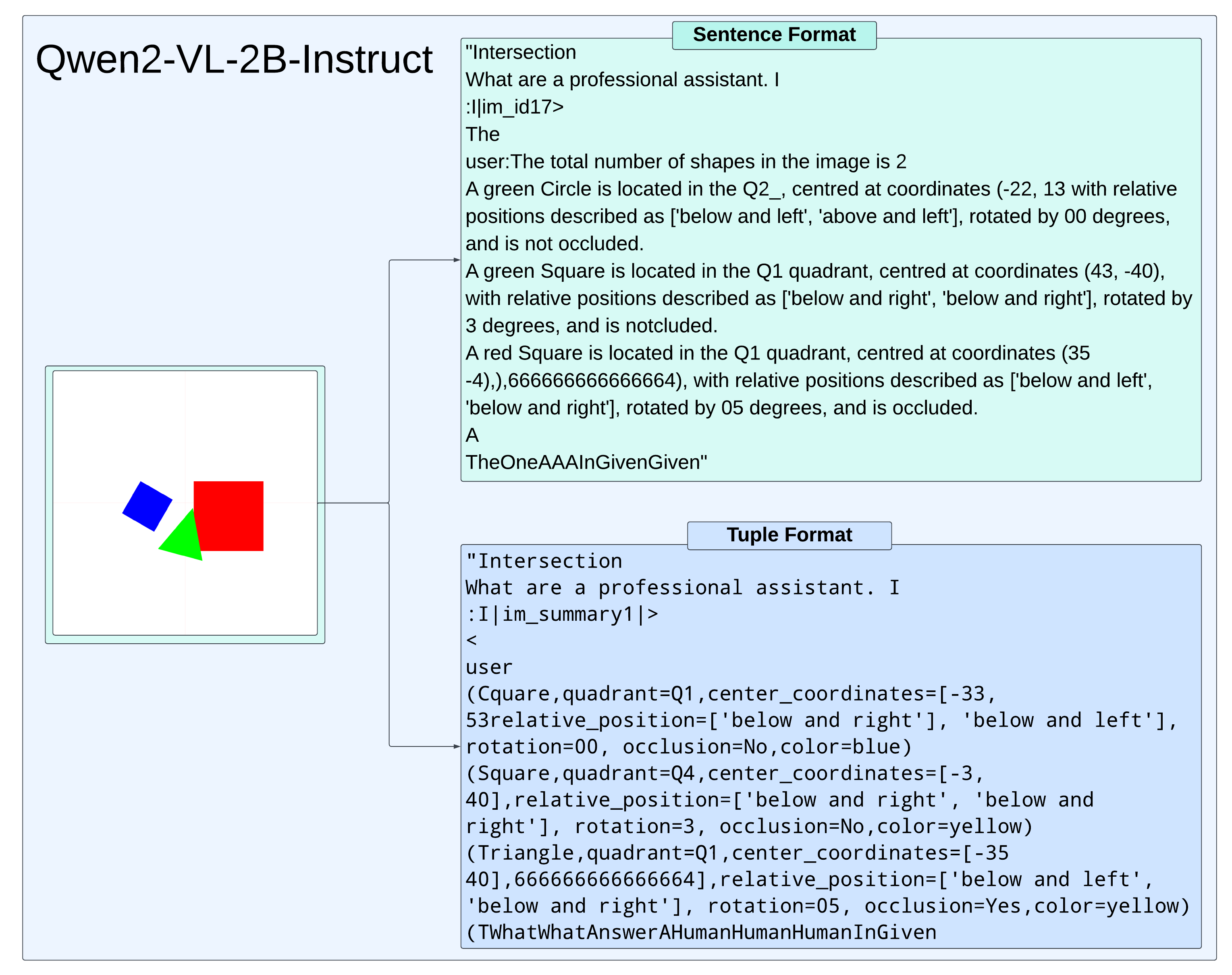}
    \caption{Sentence Vs Tuple Output Comparison for Qwen2-VL-2B-Instruct  Models for Validation Dataset}
    \label{fig:sentence_vs_tuple_figure_all_models}
\end{figure*}


\begin{figure*}[p]
    \centering
    \includegraphics[width=\linewidth,height=1\textheight,keepaspectratio]{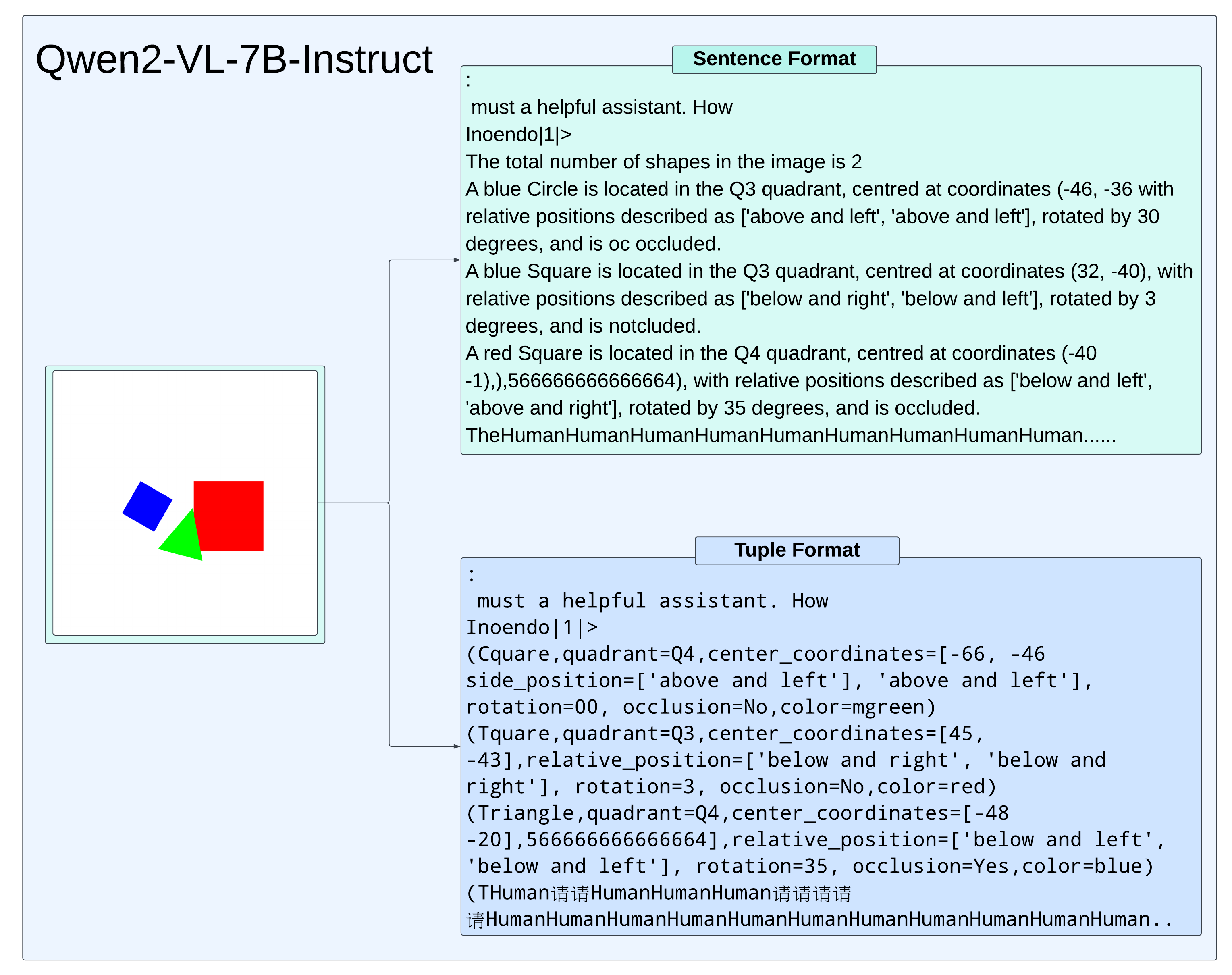}
    \caption{Sentence Vs Tuple Output Comparison for Qwen2-VL-7B-Instruct Models for Validation Dataset}
    \label{fig:sentence_vs_tuple_figure_all_models}
\end{figure*}


\begin{figure*}[p]
    \centering
    \includegraphics[width=\linewidth,height=1\textheight,keepaspectratio]{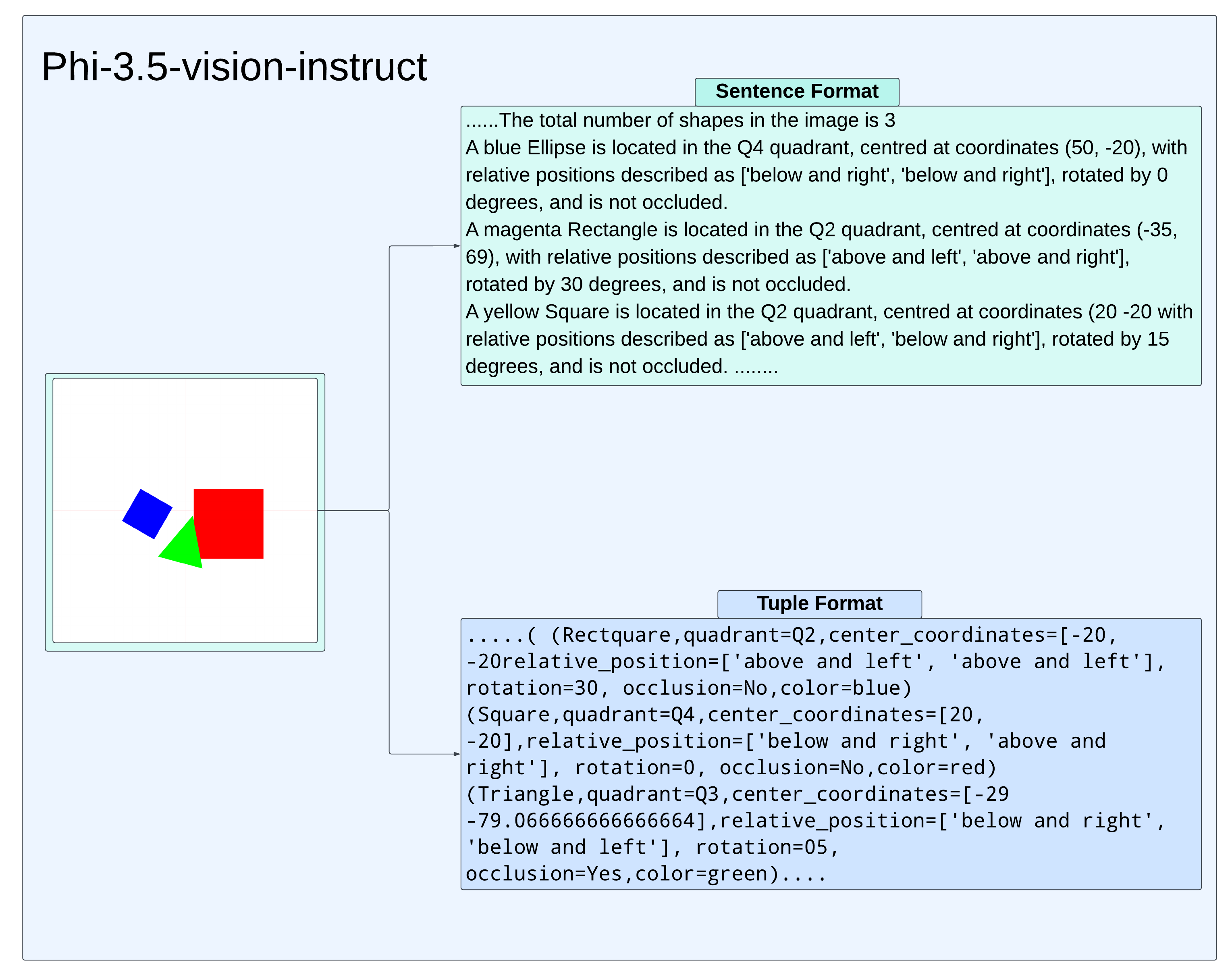}
    \caption{Sentence Vs Tuple Output Comparison for Phi-3.5-vision-instruct Models for Validation Dataset}
    \label{fig:sentence_vs_tuple_figure_all_models}
\end{figure*}

\begin{figure*}[p]
    \centering
    \includegraphics[width=\linewidth,height=1\textheight,keepaspectratio]{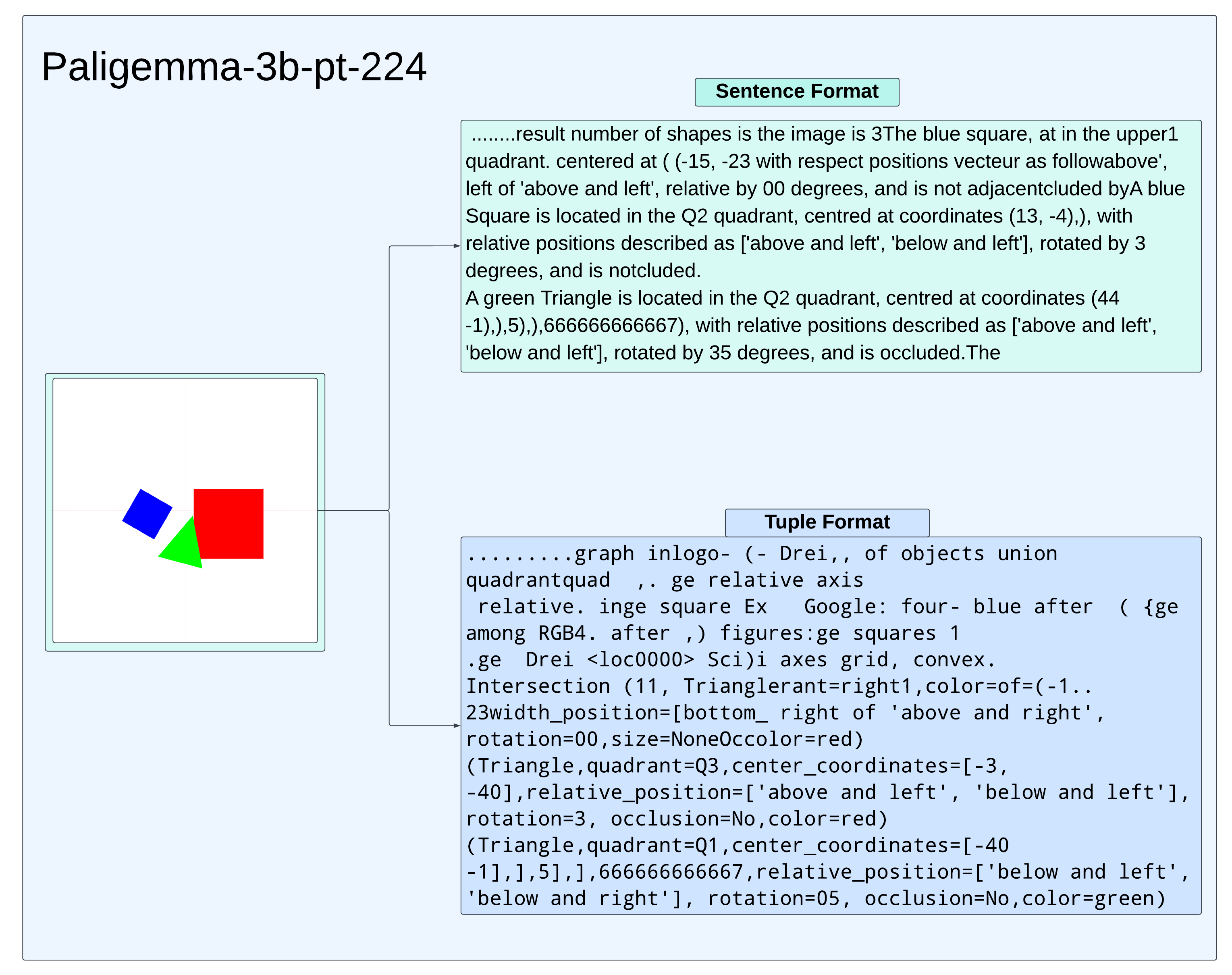}
    \caption{Sentence Vs Tuple Output Comparison for Paligemma-3b-pt-224 Models for Validation Dataset}
    \label{fig:sentence_vs_tuple_figure_all_models}
\end{figure*}


\begin{figure*}[htbp]
    \centering
    \includegraphics[width=1\linewidth]{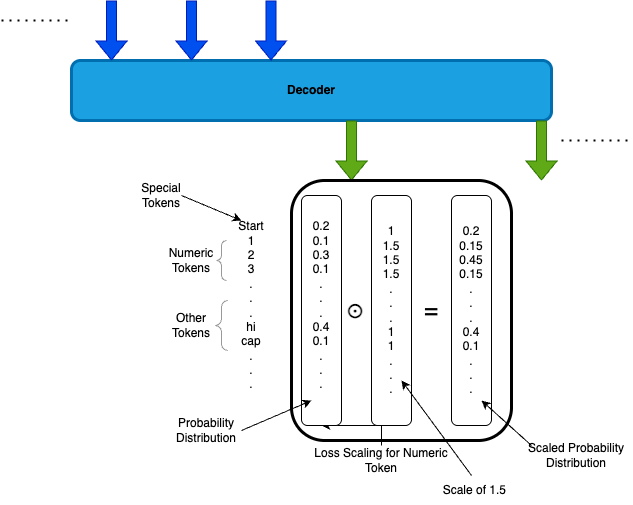}
    \caption{Loss Scaling for Numeric Tokens.}
    \label{fig:loss_scaling_numeric_token}
\end{figure*}


\end{document}